\documentclass[10pt,twocolumn,letterpaper]{article}

\usepackage{mypaper}
\usepackage{times}
\usepackage{epsfig}
\usepackage{graphicx}
\usepackage{amsmath}
\usepackage{amssymb}
\usepackage{booktabs}
\usepackage{multirow}
\usepackage{textcomp}
\usepackage{stfloats}
\usepackage{verbatim}
\usepackage{graphicx}
\usepackage{color,xcolor,colortbl}
\definecolor{lightgray}{gray}{0.95}
\definecolor{lightblue}{HTML}{95bddc}
\usepackage{bbding}
\usepackage{adjustbox}
\usepackage{makecell}
\usepackage{balance}

\newcommand{\widthscalefive}{0.28}
\newcommand{\widthscalethree}{0.4}
\usepackage{algorithmic}
\usepackage{algorithm}

\usepackage[pagebackref=true,breaklinks=true,colorlinks,bookmarks=false,citecolor=lightblue ]{hyperref}
\mypaperfinalcopy 

\ifmypaperfinal\fi

\begin{document}

\title{DiffIR: Efficient Diffusion Model for Image Restoration}

\author{%
	Bin Xia $^{1}$, Yulun Zhang $^{2}$, Shiyin Wang  $^{3}$, Yitong Wang $^{3}$, \\ Xinglong Wu $^{3}$, Yapeng Tian $^4$, Wenming Yang $^1$,  and Luc Van Gool $^2$ \\
	$^{1}$ Tsinghua University, $^2$ ETH Z\"{u}rich, $^3$  ByteDance Inc, $^4$ University of Texas at Dallas
}

\maketitle
\ifmypaperfinal\thispagestyle{empty}\fi

\begin{abstract}
\vspace{-3mm}
Diffusion model (DM) has achieved SOTA performance by modeling the image synthesis process into a sequential application of a denoising network. However, different from image synthesis, image restoration (IR) has a strong constraint to generate results in accordance with ground-truth. Thus, for IR, traditional DMs running massive iterations on a large model to estimate whole images or feature maps is inefficient. To address this issue, we propose an efficient DM for IR (DiffIR), which consists of a compact IR prior extraction network (CPEN), dynamic IR transformer (DIRformer), and denoising network. Specifically, DiffIR has two training stages: pretraining and training DM. In pretraining, we input ground-truth images into CPEN$_{S1}$ to capture a compact IR prior representation (IPR) to guide DIRformer. In the second stage, we train the DM to directly estimate the same IRP as pretrained CPEN$_{S1}$ only using LQ images. We observe that since the IPR is only a compact vector,  DiffIR can use fewer iterations than traditional DM to obtain accurate estimations and generate more stable and realistic results. Since the iterations are few, our DiffIR can adopt a joint optimization of CPEN$_{S2}$, DIRformer, and denoising network, which can further reduce the estimation error influence. We conduct extensive experiments on several IR tasks and achieve SOTA performance while consuming less computational costs. Code is available at \url{https://github.com/Zj-BinXia/DiffIR}.  
   
\end{abstract}

\vspace{-3mm}
\section{Introduction}
\vspace{-2mm}
Image Restoration (IR) is a long-standing problem due to its extensive application value and ill-posed nature. IR aims to restore a high-quality (HQ) image from its low-quality (LQ) counterpart corrupted by various degradation factors (\eg, blur, mask, downsampling). Presently, deep-learning based IR methods have achieved impressive success, as they can learn strong priors from large-scale datasets.

Recently, Diffusion Models (DMs)~\cite{DDPM}, which is built from a hierarchy of denoising autoencoders, have achieved impressive results in image synthesis~\cite{DDPM2,score-diffusion, DDPM3, DDPM4} and IR tasks (such as inpainting~\cite{repaint, LDM} and super-resolution~\cite{SR3}). Specifically, DMs are trained to iteratively denoise the image by reversing a diffusion process. DMs have shown that the principled probabilistic diffusion modeling can realize high-quality mapping from randomly sampled Gaussian noise to the complex target distribution, such as a realistic image or latent~\cite{LDM} distribution, without suffering mode-collapse and training instabilities as GANs. 

As a class of likelihood-based models, DMs require a large number of iteration steps (about $50-1000$ steps) on large denoising models to model precise details of the data, which consumes massive computational resources. Unlike the image synthesis tasks generating each pixel from scratch, IR tasks only require adding accurate details on the given LQ images. Therefore, if DMs adopt the paradigm of image synthesis for IR, it would not only waste a large number of computational resources but also be easy to generate some details that do not match given LQ images. 

\begin{figure*}[t]
	\centering
	\includegraphics[height=8cm]{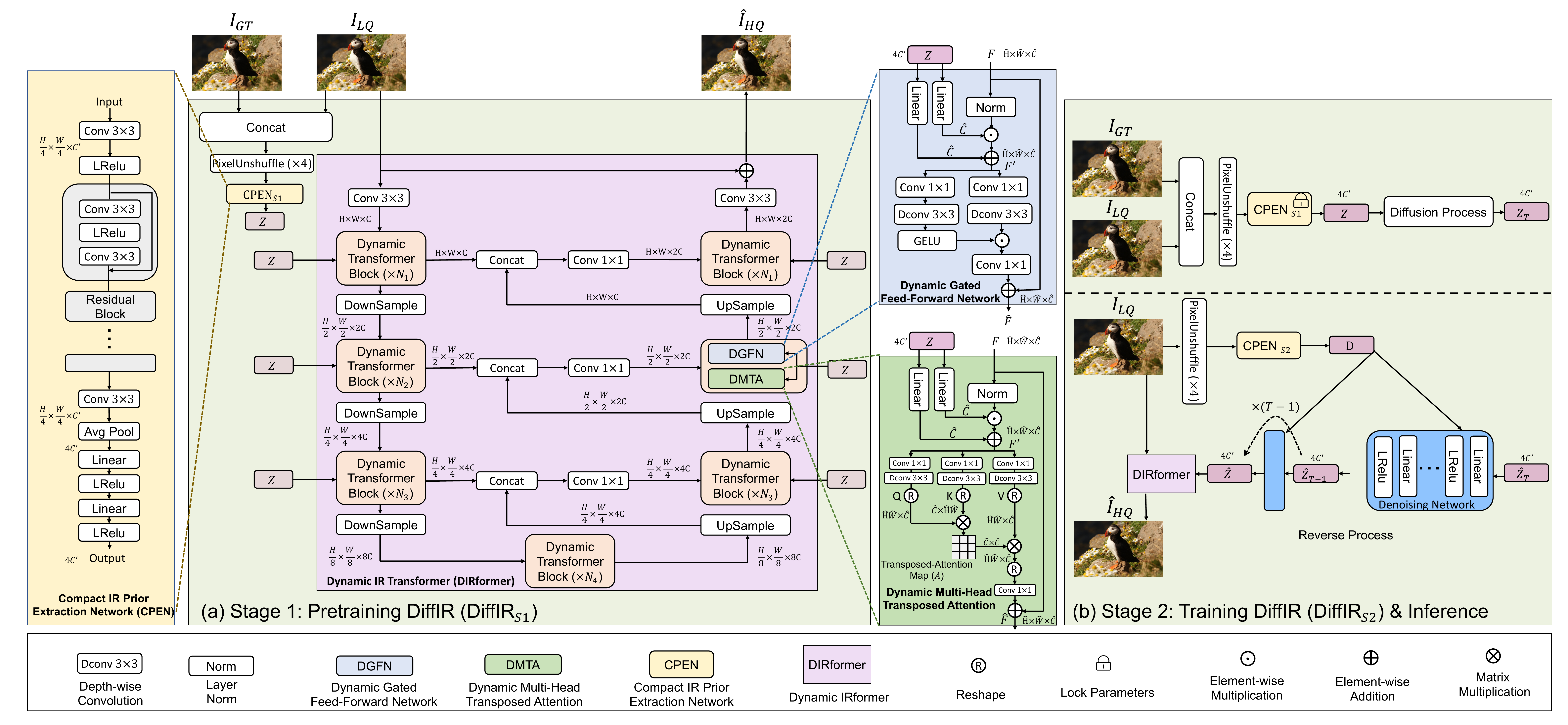}
	\caption{The overview of the proposed DiffIR, which consists of DIRformer, CPEN, and denoising network. DiffIR has two training stages: (a) In the first stage, CPEN$_{S1}$ takes the ground-truth image as input and outputs an IPR $\mathbf{Z}$ to guide DIRformer to restore images. We optimize the CPEN$_{S1}$ with DiffIR$_{S1}$ together to make DiffIR$_{S1}$ can fully use extracted IPR. (b) In the second stage, we use the strong data estimation abilities of the DM to estimate the IPR extracted by pretrained CPEN$_{S1}$. Notably, we do not input the ground-truth image into CPEN$_{S2}$ and denoising networks. In the inference stage, we only use the reverse process of DM. }
	\label{fig:method}
 \vspace{-3mm}
\end{figure*}

In this paper, we aim to design a DM-based IR network that can fully and efficiently use the powerful distribution mapping abilities of DM to restore images. To this end, we propose DiffIR. Since the transformer can model long-range pixel dependencies, we adopt the transformer blocks as our basic unit of DiffIR. We stack transformer blocks in Unet shape to form Dynamic IRformer (DIRformer) to extract and aggregate multi-level features. We train our DiffIR in two stages: \textbf{(1)} In the first stage (Fig.~\ref{fig:method} (a)), we develop a compact IR prior extraction network (CPEN) to extract a compact IR prior representation (IPR) from ground-truth images to guide the DIRformer. Besides, we develop Dynamic Gated Feed-Forward Network (DGFN) and Dynamic Multi-Head Transposed Attention (DMTA) for DIRformer to fully use the IPR. It is notable that CPEN and DIRformer are optimized together. \textbf{(2)}  In the second stage (Fig.~\ref{fig:method} (b)), we train the DM to directly estimate the accurate IPR from LQ images. Since the IPR is light and only adds details for restoration, our DM can estimate quite an accurate IPR and obtain stable visual results after several iterations. 

\begin{figure*}[t]
\scriptsize
\centering
\begin{tabular}{ccc}
    \hspace{-4mm} \includegraphics[height=0.24\textwidth]{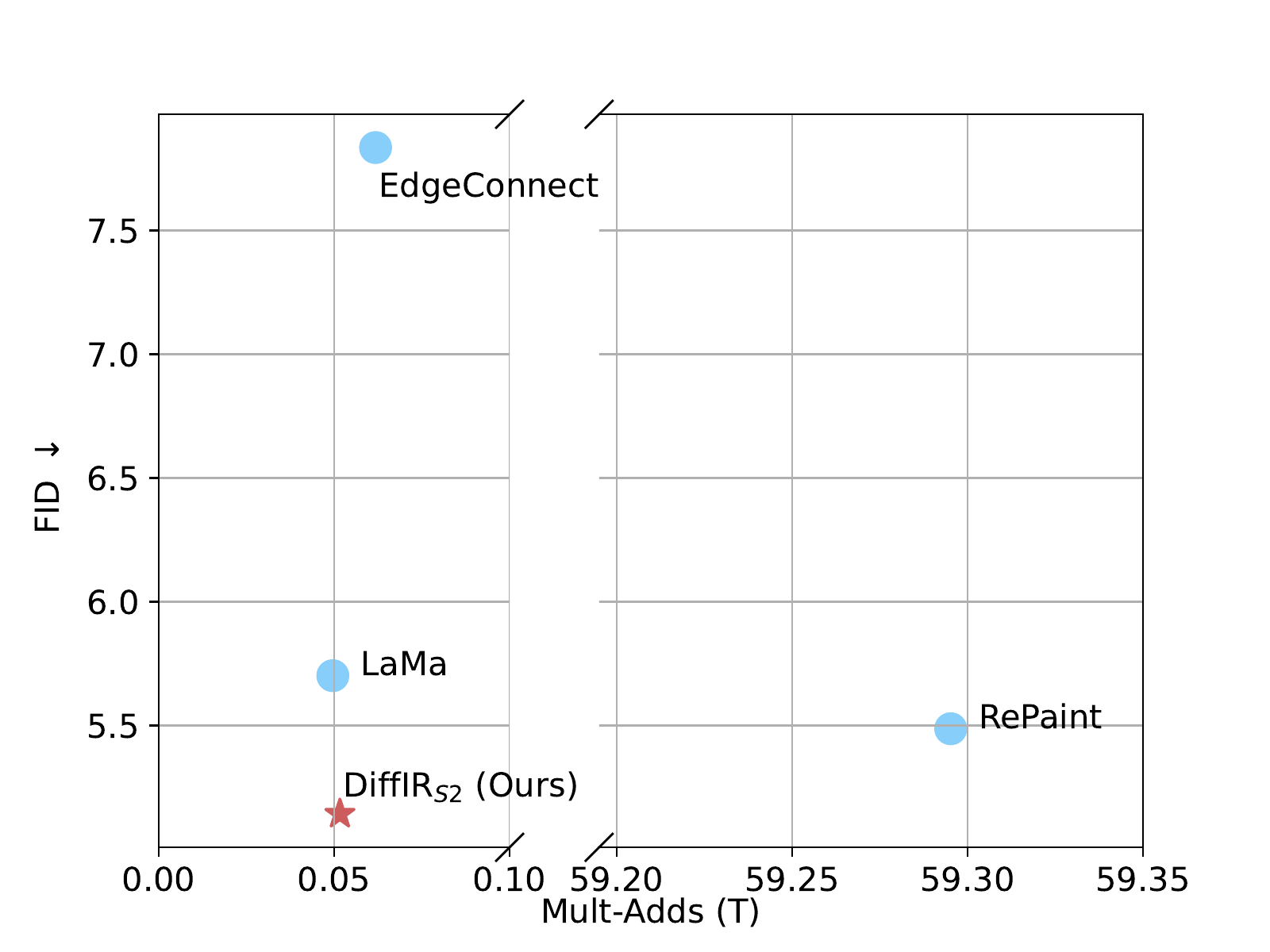}  & 
    \hspace{-6mm} \includegraphics[height=0.24\textwidth]{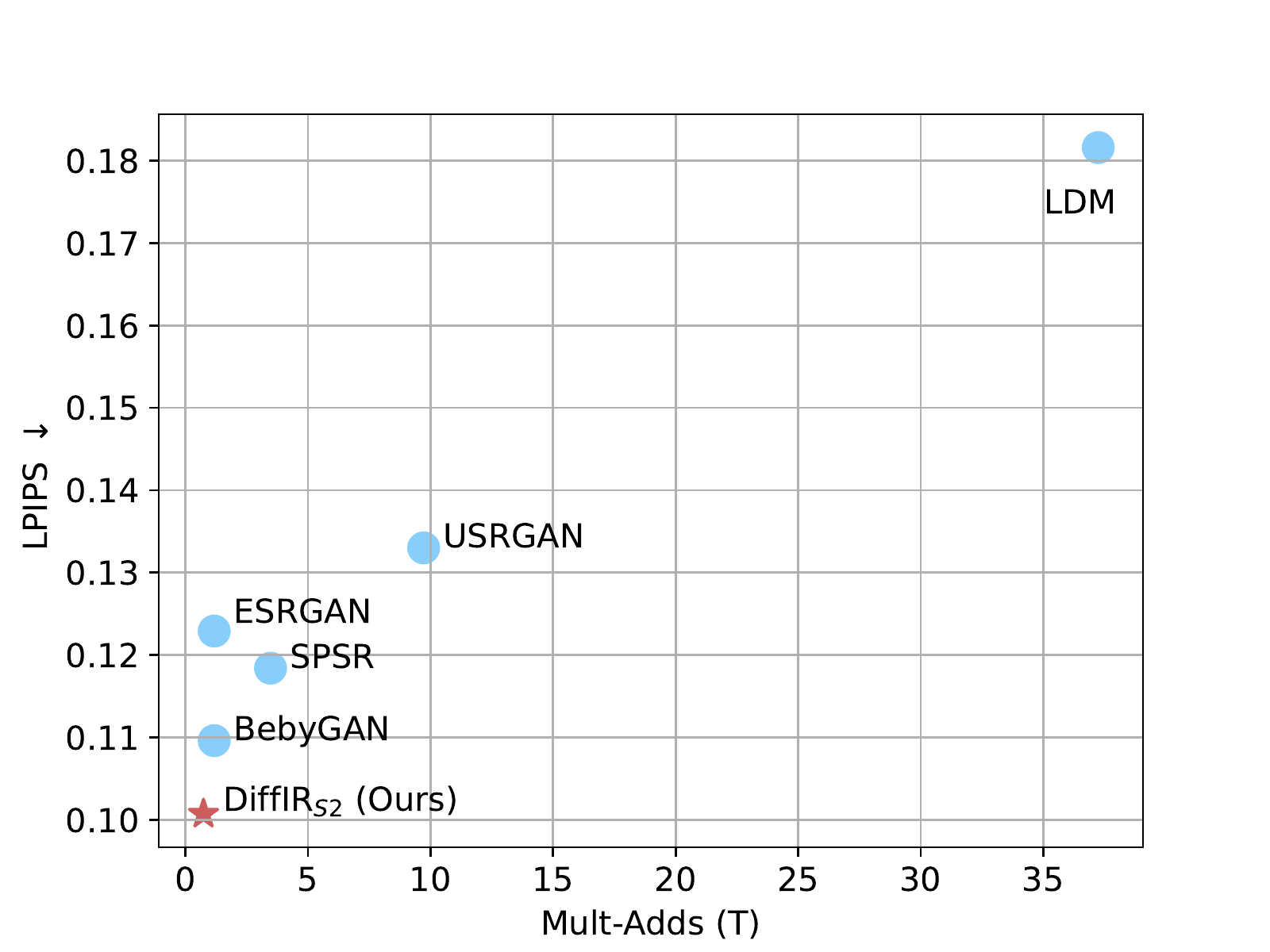}	 &
    \hspace{-6mm} \includegraphics[height=0.24\textwidth]{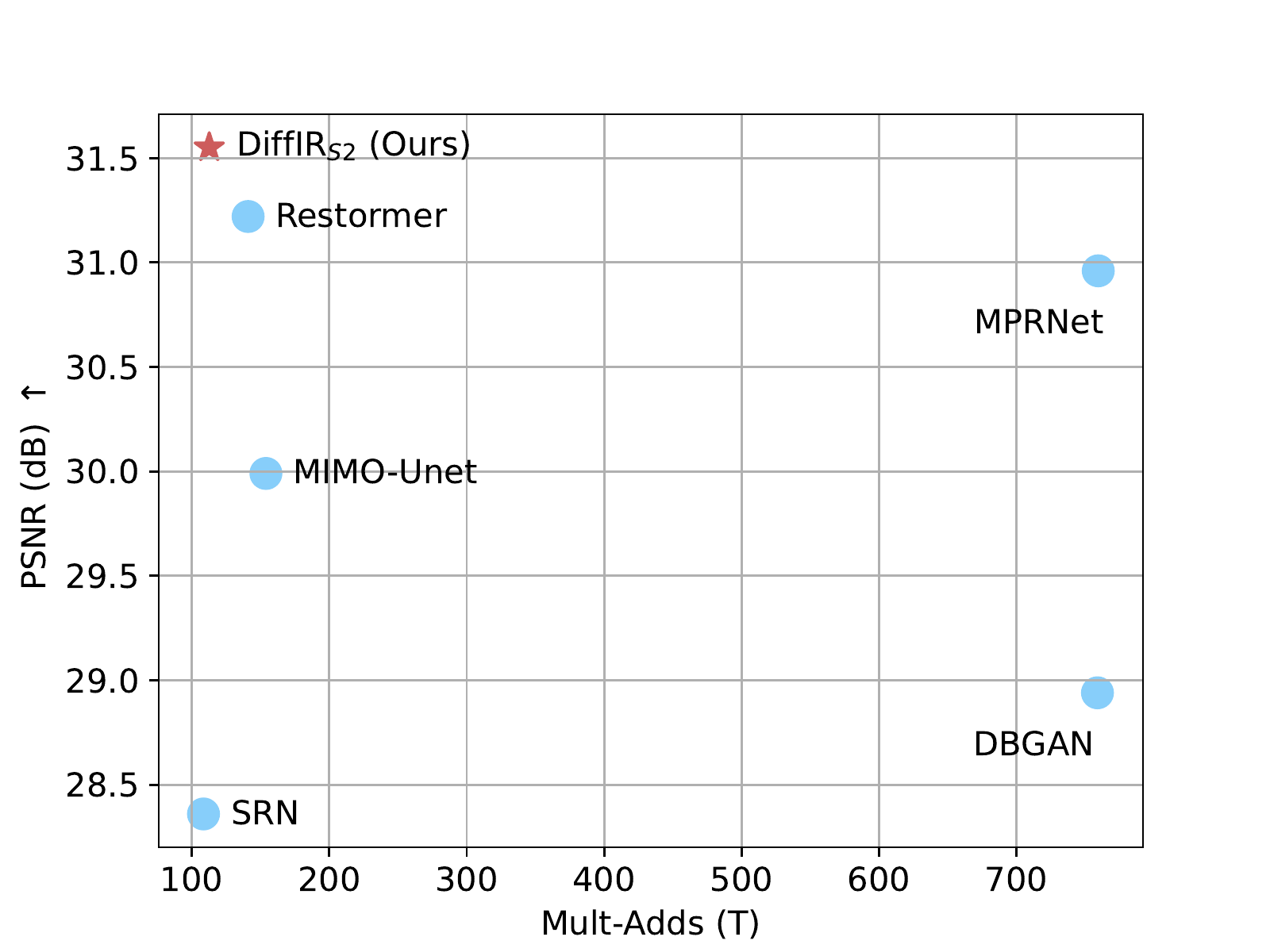}  \\

    \hspace{-5mm} \shortstack{(a) Inpainting (Tab.~\ref{tab:inpainting})}   &
    \hspace{-2mm} \shortstack{(b) Super-Resolution (Tab.~\ref{tab:SR})}   &	
    \hspace{-2mm} \shortstack{(c) Motion deblurring (Tab.~\ref{tab:deblur})}   
\end{tabular}
\label{fig:efficiency}
\vspace{1mm}
\caption{The Mult-Adds are measured on 256$\times$256 inputs. Our DiffIR achieves SOTA performance on IR tasks. Notably, LDM~\cite{LDM} and RePaint~\cite{repaint} are DM-based methods, and DiffIR is \textbf{1000$\times$ more efficient} than RePaint while achieving better performance.  
}
\vspace{-4mm}
\label{fig:head}
\end{figure*}

Apart from the above scheme and architectural novelties, we show the effectiveness of joint optimization. In the second stage, we observe that the estimated IPR may still have minor errors, which will affect the performance of the DIRformer. However, the previous DMs need many iterations, which is unavailable to optimize DM with the decoder together. Since our DiffIR requires few iterations, we can run all iterations and obtain the estimated IPR to optimize with DIRformer jointly. As shown in Fig.~\ref{fig:head}, our DiffIR achieves SOTA performance consuming much less computation than other DM-based methods (\eg, RePaint~\cite{repaint} and LDM~\cite{LDM}). In particular, DiffIR is 1000$\times$ more efficient than RePaint. Our main contributions are threefold:
 \begin{itemize}
 \item  We propose DiffIR, a strong, simple, and efficient DM-based baseline for IR. Unlike image synthesis, most pixels of input images in IR are given. Thus, we use the strong mapping abilities of DM to estimate a compact IPR to guide IR, which can improve the restoration efficiency and stability for DM in IR. 
\item  We propose DGTA and DGFN for Dynamic IRformer to fully exploit the IPR. Different from the previous latent DMs optimizing the denoising network individually, we propose joint optimization of the denoising network and decoder (\ie, DIRformer) to further improve the robustness of estimation errors.     
\item  Extensive experiments show that the proposed DiffIR can achieve SOTA performance in IR tasks while consuming much less computational resources compared with other DM-based methods.  
\end{itemize}


\vspace{-2mm}
\section{Related Work}
\vspace{-1mm}
\noindent
\textbf{Image Restoration.}
As pioneer works, SRCNN~\cite{SRCNN}, DnCNN~\cite{DnCNN}, and ARCNN~\cite{ARCNN} adopt compact CNN to achieve impressive performance on IR. After that, CNN-based methods became more popular compared with traditional IR methods. Up to now, researchers have carried out CNN's study with different perspectives and obtained more elaborate network architecture designs and learning schemes, such as residual block~\cite{VDSR,plug-denoiser,CAS-CNN}, GAN~\cite{WGAN-GP,ESRGAN,inpainting-GAN}, attention~\cite{RCAN,ENLCA,SAN,inpainting1,inpainting2,inpainting3,deepfillv2}, knowledge distillation~\cite{KDSR}, and others~\cite{restor12,restor13,restor14,restor15, AOTGAN}.

Recently, transformer, a natural language processing model, has gained much popularity in the computer
vision community. Compared with CNN, transformers can model global interactions between different regions and achieve state-of-the-art performance. Presently, the transformer has been adopted in numerous vision tasks, such as image recognition~\cite{VIT,touvron2021training}, segmentation~\cite{seg1,seg2,seg3,U-net-transformer}, object detection~\cite{detect1,detect2}, and image restoration~\cite{IPT,SWINIR,restormer,MAT,NAFNet}. 

\vspace{1mm}
\noindent
\textbf{Diffusion Models.} Diffusion Models (DMs)~\cite{DDPM2},
have achieved state-of-the-art results in density estimation~\cite{kingma2021variational} as well as in sample quality~\cite{DDPM3}. DMs adopt parameterized Markov chain to optimize the lower variational bound on the likelihood function, which can make them generate more accurate target distribution than other generative models, \ie, GAN. Recently, DM has become increasingly influential in the field of image restoration tasks, such as super-resolution~\cite{DDPM7,SR3} and inpainting~\cite{repaint,LDM,chung2022come}. SR3~\cite{SR3} and SRdiff~\cite{srdiff} introduced a DM to image super-resolution and achieved better performance than SOTA GAN-based methods. Besides, Palette~\cite{Palette} is inspired by conditional generation models~\cite{mirza2014conditional} and proposes a conditional diffusion model for IR. LDM~\cite{LDM} proposes to perform DM on latent space to improve the restoration efficiency. Furthermore, RePaint~\cite{repaint} designs an improved denoising strategy by resampling iterations in DM for inpainting. However, these DM-based IR methods directly use the paradigm of DM in image synthesis. However, most of the pixels in IR are given, and it is unnecessary to perform DM on whole images or feature maps. Our DiffIR performs DM on a compact IPR, which can make DM process more efficient and stable for IR.   

\section{Preliminaries: Diffusion Models}
In this paper, we adopt diffusion models (DMs)~\cite{DDPM2} to generate accurate IR prior representation (IPR). In the training phase, DM methods define a diffusion process that transforms an input image $x_0$ to Gaussian noise $x_T \sim \mathcal{N}(0,1)$ by $T$ iterations. Each iteration of the diffusion process can be described as follows:
\vspace{-1mm}
\begin{equation}
q\left(x_t \mid x_{t-1}\right)=\mathcal{N}\left(x_t ; \sqrt{1-\beta_t} x_{t-1}, \beta_t \mathbf{I}\right),
\label{eq:diff1}
\vspace{-1mm}
\end{equation}
where $x_t$ is the noised image at time-step $t$, $\beta_t$ is the predefined scale factor, and $\mathcal{N}$ represents the Gaussian distribution. The Eq.~\eqref{eq:diff1} can be further simplified as follows:
\begin{equation}
q\left(\mathbf{x}_t \mid \mathbf{x}_0\right)=\mathcal{N}\left(\mathbf{x}_t ; \sqrt{\bar{\alpha}_t} \mathbf{x}_0,\left(1-\bar{\alpha}_t\right) \mathbf{I}\right),
\label{eq:diff2}
\end{equation}
where $\alpha_t=1-\beta_t$, $\bar{\alpha}_t=\prod_{i=0}^t \alpha_i$.

 In the inference stage (reverse process), DM methods sample a Gaussian random noise map $x_T$ and then gradually denoise $x_T$ until it reaches a high-quality output $x_0$:
\begin{equation}
p\left(\mathbf{x}_{t-1} \mid \mathbf{x}_t, \mathbf{x}_0\right)=\mathcal{N}\left(\mathbf{x}_{t-1} ; \boldsymbol{\mu}_t\left(\mathbf{x}_t, \mathbf{x}_0\right), \sigma_t^2 \mathbf{I}\right),
\label{eq:diff3}
\end{equation}
where mean $\boldsymbol{\mu}_t\left(\mathbf{x}_t, \mathbf{x}_0\right)=\frac{1}{\sqrt{\alpha_t}}\left(\mathbf{x}_t-\epsilon \frac{1-\alpha_t}{\sqrt{1-\bar{\alpha}_t}}\right)$ and variance $\sigma_t^2=\frac{1-\bar{\alpha}_{t-1}}{1-\bar{\alpha}_t} \beta_t$. $\epsilon$ indicates the noise in $x_t$, which is the only uncertain variable in the reverse process. DMs adopt a denoising network $\epsilon_{\theta}(x_t,t)$ to estimate $\epsilon$. To train $\epsilon_{\theta}(x_t,t)$, given a clean image $x_0$, DMs randomly sample a time step $t$ and a noise $\epsilon \sim \mathcal{N}(0, \mathbf{I})$ to generate noisy images $x_t$ according to Eq.~\eqref{eq:diff2}. Then, DMs optimize the network parameters $\theta$ of $\epsilon_{\theta}$ following \cite{DDPM2}:
\begin{equation}
\nabla_{\boldsymbol{\theta}}\left\|\epsilon-\epsilon_{\boldsymbol{\theta}}\left(\sqrt{\bar{\alpha}_t} \mathrm{x}_0+\epsilon \sqrt{1-\bar{\alpha}_t}, t\right)\right\|_2^2.
\label{eq:diff4}
\end{equation}

\section{Methodology}
Traditional DMs~\cite{DDPM,LDM,repaint} require a large number of iterations, computational resources, and model parameters to generate accurate and realistic images or latent feature maps. Although DMs achieve impressive performance in generating images from scratch (image synthesis), it is a waste of computational resources to directly apply the DM paradigm of image synthesis to IR. Since most pixels and information in IR are given, performing DMs on whole images or feature maps not only spends a lot of iterations and computation but also is easy to generate more artifacts. Overall, DMs have strong data estimation ability, but applying the existing DM paradigm in image synthesis to IR is inefficient. To address the issue, we propose an efficient DM for IR (\ie, DiffIR), which adopts DM to estimate a compact IPR to guide the network to restore images. Since the IPR is quite light, the model size and iteration of DiffIR can be largely reduced to generate more accurate estimations compared with traditional DM.

In this section, we present our DiffIR. As shown
in Fig.~\ref{fig:method}, DiffIR mainly consists of a compact IR prior extraction network (CPEN), dynamic IRformer (DIRformer), and denoising network. We train DiffIR in two stages, including pretraining DiffIR and training the diffusion model. In the following sections, we first introduce the pretraining DiffIR in Sec.~\ref{sec:pretrain}. Then, we provide the details of the training efficient DM for DiffIR in Sec.~\ref{sec:diffusion}.

\vspace{-1mm}
\subsection{Pretrain DiffIR}
\label{sec:pretrain}
Before introducing pretraining DiffIR, we would like to introduce two networks in the first stage, including a compact IR prior extraction network (CPEN) and a dynamic IRformer (DIRformer). The structure of CPEN is shown in Fig.~\ref{fig:method} yellow box, which is mainly stacked with residual blocks and linear layers to extract the compact IR prior representation (IPR). After that, DIRformer can use the extracted IPR to restore LQ images. The structure of the DIRformer is shown in Fig.~\ref{fig:method} pink box, which is stacked with dynamic transformer blocks in the Unet shape. The dynamic transformer blocks consist of dynamic multi-head transposed attention (DMTA, Fig.~\ref{fig:method} green box) and dynamic gated feed-forward network (DGFN, Fig.~\ref{fig:method} nattier blue box), which can use IPR as dynamic modulation parameters to add restoration details into feature maps. 

In the pretraining (Fig.~\ref{fig:method}~(a)), we train CPEN$_{S1}$ and DIRformer together. Specifically, we first concatenate ground-truth and LQ images together and use the PixelUnshuffle operation to downsample them to obtain the input for CPEN$_{S1}$. Then, CPEN$_{S1}$ extract the IPR $\mathbf{Z}\in \mathbb{R}^{4C^{\prime}}$  as:
\begin{equation}
\mathbf{Z}=\operatorname{CPEN_{S1}}(\operatorname{PixelUnshuffle}(\operatorname{Concat}(I_{GT},I_{LQ}))). 
\end{equation}
Then IPR $\mathbf{Z}$ is sent into DGFN and DMTA of DIRformer as dynamic modulation parameters to guide restoration:
\begin{equation}
\mathbf{F^{\prime}}=W_{l}^{1}\mathbf{Z}\odot\operatorname{Norm}(\mathbf{F}) + W_{l}^{2}\mathbf{Z},
\end{equation}
where $\odot$ indicates element-wise multiplication, $\operatorname{Norm}$ denotes layer normalization~\cite{layernorm}, $W_{l}$ represents linear layer, $\mathbf{F}$ and $\mathbf{F^{\prime}} \in\mathbb{R}^{\hat{H}\times\hat{W}\times\hat{C}}$ are input and output feature maps respectively, and $W_{l}^{1}\mathbf{Z}, W_{l}^{2}\mathbf{Z}\in\mathbb{R}^{\hat{C}}$.

Then, we aggregate global spatial information in DMTA. Specifically,  $\mathbf{F^{\prime}}$ is projected into query $\mathbf{Q}=W_d^Q W_c^Q \mathbf{F^{\prime}}$, key $\mathbf{K}=W_d^K W_c^K \mathbf{F^{\prime}}$, and value $\mathbf{V}=W_d^V W_c^V \mathbf{F^{\prime}}$, where
$W_c$ is the $1\times1$ point-wise convolution and $W_d$ is the $3\times3$ depth-wise convolution. 
Next, we reshape the query $\hat{\mathbf{Q}} \in \mathbb{R}^{\hat{H} \hat{W} \times \hat{C}}$, key $\hat{\mathbf{K}} \in \mathbb{R}^{\hat{C}\times\hat{H} \hat{W}}$, and value $\hat{\mathbf{V}} \in \mathbb{R}^{\hat{H} \hat{W} \times \hat{C}}$. After that, we perform dot-product between $\hat{\mathbf{Q}}$ and $\hat{\mathbf{K}}$ generates
a transposed-attention map $\mathbf{A}$ of size $\mathbb{R}^{\hat{C} \times \hat{C}}$, which is more efficient than regular attention map of size $\mathbb{R}^{\hat{H} \hat{W} \times \hat{H} \hat{W}}$. The overall process of DMTA can be described as follows:
\begin{equation}
\vspace{-1mm}
\hat{\mathbf{F}}=W_c\hat{\mathbf{V}} \cdot \operatorname{Softmax}(\hat{\mathbf{K}} \cdot \hat{\mathbf{Q}} / \gamma)+\mathbf{F},
\end{equation}
where $\gamma$ is a learnable scaling parameter. As conventional multi-head self attention~\cite{VIT,IPT} did, we separate channels to multi-head and calculate attention maps.

Next, in DGFN, we aggregate local features. We use $1\times1$ Conv to aggregate information from different channels and adopt $3\times3$ depth-wise Conv to aggregate information from spatially neighboring pixels. Besides, we adopt the gating mechanism to enhance information encoding. The overall process of DGFN is defined as:
\begin{equation}
\mathbf{\hat{F}}=\operatorname{GELU}\left(W_d^1 W_c^1\mathbf{F^{\prime}}\right) \odot W_d^2 W_c^2\mathbf{F^{\prime}}+\mathbf{F}.
\end{equation}

We train CPEN$_{S1}$ and DIRformer together, which can make DIRformer fully use the IPR extracted by CPEN$_{S1}$ for restoration. The training loss is defined as follows:
\vspace{-2mm}
\begin{equation}
L_{rec}= \left\|I_{GT}-\hat{I}_{HQ}\right\|_{1},
\vspace{-2mm}
\label{eq:rec}
\end{equation}
where $I_{GT}$ and $\hat{I}_{HQ}$ are the ground-truth and restored HQ images, respectively. $\|\cdot\|_{1}$ denotes the $L_{1}$ norm. If some works emphasize visual quality, such as inpainting and SISR, we can further add perceptual loss and adversarial loss. More details are provided in supplementary materials.

\subsection{Diffusion Models for Image Restoration}
\label{sec:diffusion}
In the second stage (Fig.~\ref{fig:method}~(b)), we exploit the strong data estimation ability of the DM to estimate IPR. Specifically, we use the pretrained CPEN$_{S1}$ to capture the IPR $\mathbf{Z}\in \mathbb{R}^{4C^{\prime}}$. After that, we apply the diffusion process on $\mathbf{Z}$ to sample $\mathbf{Z}_T\in \mathbb{R}^{4C^{\prime}}$, which can be described as: 
\vspace{-1mm}
\begin{equation}
q\left(\mathbf{Z}_T \mid \mathbf{Z}\right)=\mathcal{N}\left(\mathbf{Z}_T; \sqrt{\bar{\alpha}_T} \mathbf{Z},\left(1-\bar{\alpha}_T\right) \mathbf{I}\right),
\label{eq:mydiff1}
\end{equation}
where $T$ is the total number of iterations, $\bar{\alpha}$ and $\alpha$ are defined in Eqs.~\eqref{eq:diff1} and ~\eqref{eq:diff2} (\ie,  $\bar{\alpha}_T=\prod_{i=0}^T \alpha_i$).

In the reverse process, since IPR is compact, DiffIR$_{S2}$ can use much fewer iterations and smaller model size to obtain quite good estimations than traditional DMs~\cite{LDM, repaint}. Since traditional DMs have huge computational costs in iterations, they have to randomly sample a time-step $t\in[1,T]$ and merely optimize the denoising network at that time step (Eqs.~\eqref{eq:diff1},~\eqref{eq:diff2},~\eqref{eq:diff3}, and~\eqref{eq:diff4}). The lack of joint training of the denoising network and decoder (\ie, DIRformer) means the minor error of estimations caused by the denoising network would make the DIRformer cannot achieve its potential. By contrast, DiffIR starts from $T$-th time step (Eq.~\eqref{eq:mydiff1}) and runs all denoising iterations (Eq.~\eqref{eq:mydiff2}) to obtain $\mathbf{\hat{Z}}$ and send it to DIRformer for joint optimization.  
\vspace{-1mm}
\begin{equation}
\mathbf{\hat{Z}}_{t-1}=\frac{1}{\sqrt{\alpha_t}}\left(\mathbf{\hat{Z}}_t-\epsilon \frac{1-\alpha_t}{\sqrt{1-\bar{\alpha}_t}}\right),
\label{eq:mydiff2}
\vspace{-1mm}
\end{equation}
where $\epsilon$ indicates the same noise, and we use the CPEN$_{S2}$ and denoising network to predict noise as Eq.~\eqref{eq:diff3}. It is notable that, different from traditional DMs in Eq.~\eqref{eq:diff3}, our DiffIR$_{S2}$ delete the variance estimation and find it helpful for accurate IPR estimation and better performance (Sec.~\ref{sec:ablation}). 

In the reverse process of DM, we first use CPEN$_{S2}$ to obtain a conditional vector $\mathbf{D} \in\mathbb{R}^{4C^{\prime}}$ from LQ images:
\vspace{-1mm}
\begin{equation}
\vspace{-1mm}
\mathbf{D}=\operatorname{CPEN_{S2}}(\operatorname{PixelUnshuffle}(I_{LQ})),
\end{equation}
where CPEN$_{S2}$ has the same structure as CPEN$_{S1}$ except the input dimension of the first convolution. Then, we use the denoising network $\epsilon_{\theta}$ to estimate noise in each time step $t$ as  $\epsilon_{\theta}(\operatorname{Concat}(\mathbf{\hat{Z}}_t,t,\mathbf{D}))$. The estimated noise is substituted into Eq.~\eqref{eq:mydiff2} to obtain $\mathbf{\hat{Z}}_{t-1}$ to start the next iteration. 

Then, after $T$ times iterations, we obtain the final estimated IPR $\mathbf{\hat{Z}} \in\mathbb{R}^{4C^{\prime}}$. We joint train CPEN$_{S2}$, denoising network, and DIRformer using $\mathcal{L}_{all}$:
\begin{equation}
\label{eq:diff}
\vspace{-1mm}
\mathcal{L}_{diff}=\frac{1}{4C^{\prime}}\sum_{i=1}^{4C^\prime}\left|\hat{\mathbf{Z}}(i)-\mathbf{Z}(i)\right |, 
\mathcal{L}_{all}=\mathcal{L}_{rec}+\mathcal{L}_{diff},
\end{equation}
where we can further add perceptual loss and adversarial loss in $\mathcal{L}_{all}$ for better visual quality as Eq.~\eqref{eq:rec}.

In the inference stage, we only use the reverse diffusion process (the bottom part of Fig.~\ref{fig:method} (b)). CPEN$_{S2}$ extracts a conditional vector $\mathbf{D}$ from LQ images, and we randomly sample a Gaussian noise $\mathbf{\hat{Z}}_T$. Denoising network utilizes the $\mathbf{\hat{Z}}_T$ and $\mathbf{D}$ to estimate IPR $\mathbf{\hat{Z}}$ after $T$ iterations. After that, DIRformer exploits the IPR to restore LQ images.   

\begin{table*}[t]
  \centering
  \caption{Quantitative comparison (FID/LPIPS) for \textbf{inpainting} on benchmark datasets. Best and second best performance are marked in bold and underlined, respectively. The bottom three methods marked in gray adopt the diffusion model.}
  \resizebox{1\linewidth}{!}{
    \begin{tabular}{l|c|cccc|cccc}
    \toprule[0.2em]
    \multirow{3}[5]{*}{\textbf{Method}} & \multirow{3}[5]{*}{\textbf{\#Params (M)}} & \multicolumn{4}{c|}{\textbf{Places~\cite{places2} (512$\times$512)}} & \multicolumn{4}{c}{\textbf{CelebA-HQ~\cite{celeba} (256$\times$256)}} \\
\cmidrule{3-10}          &       & \multicolumn{2}{c}{Narrow Masks} & \multicolumn{2}{c|}{Wide Masks} & \multicolumn{2}{c}{Narrow Masks} & \multicolumn{2}{c}{Wide Masks} \\
\cmidrule{3-10}          &       & FID $\downarrow$  & LPIPS $\downarrow$ & FID $\downarrow$  & LPIPS $\downarrow$& FID  $\downarrow$ & LPIPS $\downarrow$& FID $\downarrow$  & LPIPS $\downarrow$\\
\midrule
    EdgeConnect~\cite{edgeconnect} & 22    & 1.3421 & 0.1106 & 8.4866 & 0.1594 & 6.9566 & 0.0922 & 7.8346 & 0.1149 \\
    ICT~\cite{ICT}   & 150   & -     & -     & -     & -     &   8.4977	& 0.0982 & 9.8794 &	0.1196\\
    LaMa~\cite{LaMa}  & 27    & \underline{0.6340} & \underline{0.0898} & 2.2494 & \underline{0.1339} & 5.3889 & \underline{0.0806} & 5.7023 & \underline{0.0951} \\
    \rowcolor{lightgray}
    LDM~\cite{LDM}   & 215   & -     & -     & \underline{2.1500}  & 0.1440 & -     & -     & -     & - \\
    \rowcolor{lightgray}
    RePaint~\cite{repaint} & 607   & -     & -     & -     & -     &  \underline{4.7395}	& 0.0890	& \underline{5.4881} & 0.1094  \\
    \rowcolor{lightgray}
    DiffIR$_{S2}$ (Ours) & 26    & \textbf{0.4913} & \textbf{0.0758} & \textbf{1.9788} & \textbf{0.1306} & \textbf{4.5967} & \textbf{0.0769} & \textbf{5.1440} & \textbf{0.0918} \\
    \bottomrule[0.2em]
    \end{tabular}%
    }
    \vspace{-3mm}
  \label{tab:inpainting}%
\end{table*}%

\begin{figure*}[t]
    \newlength\fsdurthree
    \setlength{\fsdurthree}{0mm}
    \Huge
    \centering
   \resizebox{1\linewidth}{!}{
            \begin{adjustbox}{valign=t}
                \begin{tabular}{cccccc}
                    \includegraphics[width=\widthscalethree \textwidth]{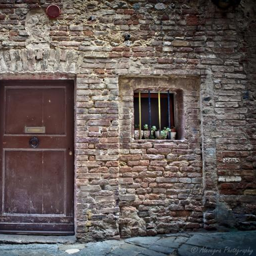} \hspace{\fsdurthree} &
                    \includegraphics[width=\widthscalethree  \textwidth]{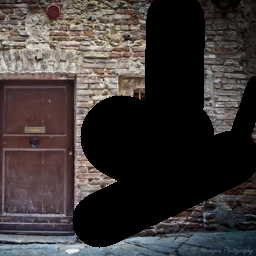} \hspace{\fsdurthree} &
                    \includegraphics[width=\widthscalethree  \textwidth]{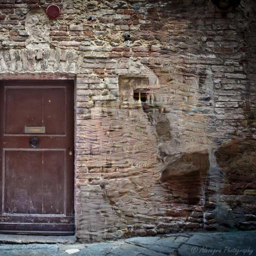} \hspace{\fsdurthree} &
                    \includegraphics[width=\widthscalethree  \textwidth]{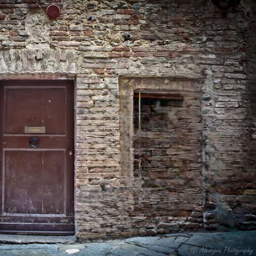} 
                    \hspace{\fsdurthree} &
                    \includegraphics[width=\widthscalethree  \textwidth]{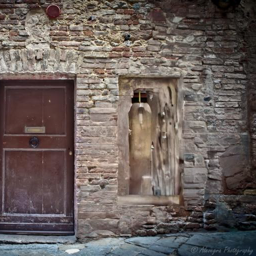} \hspace{\fsdurthree} &
                    \includegraphics[width=\widthscalethree  \textwidth]{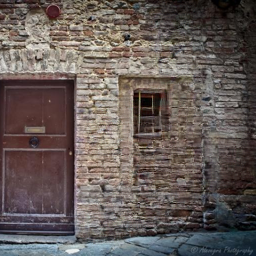} \hspace{\fsdurthree}  
                    \\
                    \includegraphics[width=\widthscalethree  \textwidth]{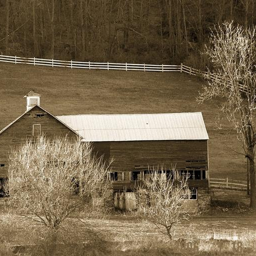} \hspace{\fsdurthree} &
                    \includegraphics[width=\widthscalethree  \textwidth]{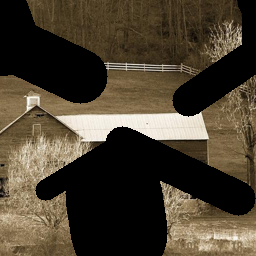} \hspace{\fsdurthree} &
                    \includegraphics[width=\widthscalethree  \textwidth]{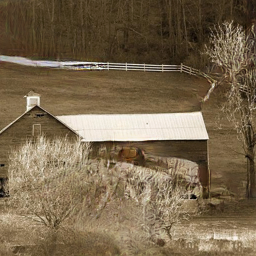} \hspace{\fsdurthree} &
                    \includegraphics[width=\widthscalethree  \textwidth]{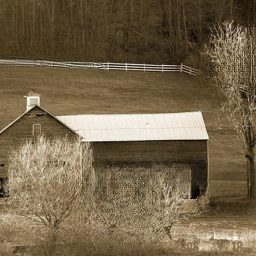} 
                    \hspace{\fsdurthree} &
                    \includegraphics[width=\widthscalethree  \textwidth]{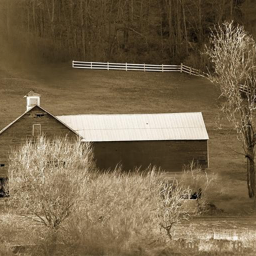} \hspace{\fsdurthree} &
                    \includegraphics[width=\widthscalethree  \textwidth]{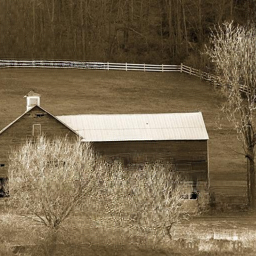} \hspace{\fsdurthree}  
                    \\
                    \includegraphics[width=\widthscalethree  \textwidth]{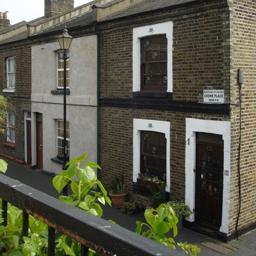} \hspace{\fsdurthree} &
                    \includegraphics[width=\widthscalethree  \textwidth]{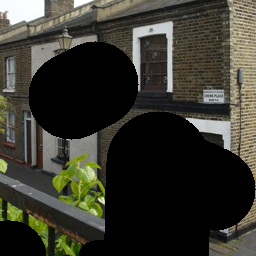} \hspace{\fsdurthree} &
                    \includegraphics[width=\widthscalethree  \textwidth]{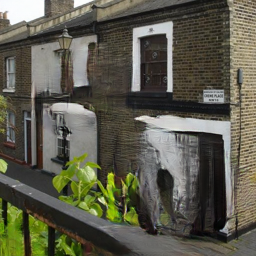} \hspace{\fsdurthree} &
                    \includegraphics[width=\widthscalethree  \textwidth]{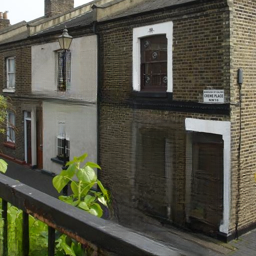} 
                    \hspace{\fsdurthree} &
                    \includegraphics[width=\widthscalethree  \textwidth]{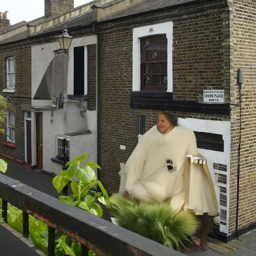} \hspace{\fsdurthree} &
                    \includegraphics[width=\widthscalethree  \textwidth]{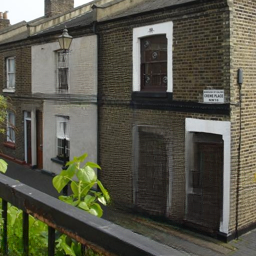} \hspace{\fsdurthree}  
                    \\
                    HQ \hspace{\fsdurthree} &
                    \makecell{LQ} \hspace{\fsdurthree} &
                    \makecell{ICT~\cite{ICT}} \hspace{\fsdurthree} &
                    LaMa~\cite{LaMa} \hspace{\fsdurthree} &
                    RePaint~\cite{repaint} \hspace{\fsdurthree} &
                    \makecell{DiffIR$_{S2}$ (Ours)} \hspace{\fsdurthree} 
                \end{tabular}
            \end{adjustbox}

    }
    \caption{Visual comparison of \textbf{ inpainting} methods. Zoom-in for better details.}
    \vspace{-3mm}
    \label{fig:inpainting_show}
\end{figure*}

\vspace{-1mm}
\section{Experiments}
\vspace{-1mm}
\subsection{Experiment Settings}
 We apply our method to three typical IR tasks separately: (a) inpainting, (b) image super-resolution (SR), (c) single-image motion deblurring. Our DiffIR adopts a 4-level encoder-decoder structure. From level-$1$ to level-$4$, the attention heads in DMTA are $[1,2,4,8]$, and the number of channels is $[48,96,192,384]$. Additionally, in all IR tasks, we tune the number of dynamic transformer blocks in DIRformer to compare DiffIR with the SOTA methods in similar parameters and computational costs. Specifically, from level-1 to level-4, we set the number of dynamic transformer blocks to $[1,1,1,9]$, $[13,1,1,1]$,  and $[3,5,6,6]$ for inpainting, SR, and deblurring, respectively. In addition, following previous works~\cite{repaint, LDM}, we introduce adversarial loss and perceptual loss for inpainting and SR. The number of channels $C^{\prime} $ of CPEN is set to $64$. 
 
In training the diffusion model, total timesteps $T$ are set to $4$, and $\beta_t$ in Eq.~\eqref{eq:mydiff2} ($\alpha_t=1-\beta_t$) linearly increase from $\beta_1=0.1$ to $\beta_T=0.99$. We train models with Adam optimizer ($\beta_{1}=0.9$, $\beta_{2}=0.99$). More details are presented in the supplementary material.

\begin{table*}[htbp]
  \centering
  \caption{Quantitative comparison ( LPIPS/DISTS). for \textbf{Single image super-resolution} on benchmark datasets. Best and second best performance are marked in bold and underlined, respectively. The bottom two methods marked in gray adopt the diffusion model.}
   \resizebox{1\linewidth}{!}{
    \begin{tabular}{l|cccccccccc}
    \toprule[0.2em]
    \multirow{2}[2]{*}{\textbf{Method}} & \multicolumn{2}{c}{\textbf{Set14}~\cite{Set14}} & \multicolumn{2}{c}{\textbf{Urban100}~\cite{Urban100}} & \multicolumn{2}{c}{\textbf{Manga109}~\cite{Manga109}} & \multicolumn{2}{c}{\textbf{General100}~\cite{general100}} & \multicolumn{2}{c}{\textbf{DIV2K100}~\cite{DIV2K}} \\
          & PSNR $\uparrow$& LPIPS $\downarrow$& PSNR $\uparrow$& LPIPS $\downarrow$& PSNR $\uparrow$& LPIPS $\downarrow$& PSNR $\uparrow$& LPIPS $\downarrow$& PSNR $\uparrow$& LPIPS $\downarrow$\\
    \midrule
    SFTGAN~\cite{SFTGAN} &    26.74  & 0.1313  & 24.34  & 0.1343  & 28.17  & 0.0716  & 29.16  & 0.0947  & 28.09  & 0.1331  \\
    SRGAN~\cite{SRGAN} &     26.84  & 0.1327  & 24.41  & 0.1439  & 28.11  & 0.0707  & 29.33  & 0.0964  & 28.17  & 0.1257  \\
    ESRGAN~\cite{ESRGAN} &    26.59  & 0.1241  & 24.37  & 0.1229  & 28.41  & 0.0649  & 29.43  & 0.0879  & 28.18  & 0.1154  \\
    USRGAN~\cite{USRGAN} &     \underline{27.41}  & 0.1347  & 24.89  & 0.1330  & 28.75  & 0.0630  & \underline{30.00}  & 0.0937  & \underline{28.79}  & 0.1325  \\
    SPSR~\cite{SPSR}  & 26.86  & 0.1207  & 24.80  & 0.1184  & 28.56  & 0.0672  & 29.42  & 0.0862  & 28.18  & 0.1099  \\
    BebyGAN~\cite{BebyGAN} & 27.09  & \underline{0.1157}  & \underline{25.23}  & \underline{0.1096}  & \underline{29.19}  & \underline{0.0529}  & 29.95  & \underline{0.0778}  & 28.62  & \underline{0.1022}  \\
    \rowcolor{lightgray}
    LDM~\cite{LDM}   & 25.62  & 0.2034  & 23.36  & 0.1816  & 25.87  & 0.1321  & 27.17  & 0.1655  & 26.66  & 0.1939  \\
    \rowcolor{lightgray}
    SRdiff~\cite{srdiff}   & 27.14	& 0.1450 &	25.12 &	0.1379 &	28.67	& 0.0665 &	29.83 &	0.1009 &	28.58 &	0.1293 \\
    \rowcolor{lightgray}
    DiffIR$_{S2}$ (Ours) & \textbf{27.73}  & \textbf{0.1117}  & \textbf{26.05}  & \textbf{0.1007}  & \textbf{30.32}  & \textbf{0.0463}  & \textbf{30.58}  & \textbf{0.0717}  & \textbf{29.13}  & \textbf{0.0871}  \\
    \bottomrule[0.2em]
    \end{tabular}%
    }
  \label{tab:SR}%
  \vspace{-2mm}
\end{table*}%

\begin{figure*}[t]

    \setlength{\fsdurthree}{0mm}
    \LARGE
    \centering
   \resizebox{1\linewidth}{!}{
        \begin{tabular}{cc}
            \begin{adjustbox}{valign=t}
                \Large
                \begin{tabular}{c}
                    \includegraphics[height=0.44\textwidth]{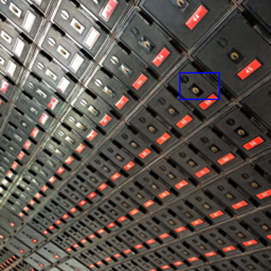} 
                \end{tabular}
                
            \end{adjustbox}
            
            \begin{adjustbox}{valign=t}
                \begin{tabular}{cccc}
                    \includegraphics[width=\widthscalefive \textwidth]{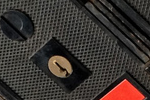} \hspace{\fsdurthree} &
                    \includegraphics[width=\widthscalefive \textwidth]{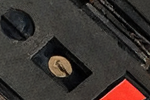} \hspace{\fsdurthree} &
                    \includegraphics[width=\widthscalefive \textwidth]{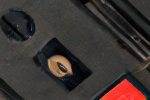} \hspace{\fsdurthree} 
                    \\
                    HQ \hspace{\fsdurthree} &
                    \makecell{BebyGAN} \hspace{\fsdurthree} &
                    \makecell{LDM} \hspace{\fsdurthree} 
                    \\
                    \includegraphics[width=\widthscalefive \textwidth]{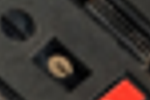} 
                    \hspace{\fsdurthree} &
                    \includegraphics[width=\widthscalefive \textwidth]{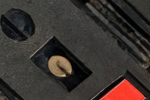} \hspace{\fsdurthree} &
                    \includegraphics[width=\widthscalefive \textwidth]{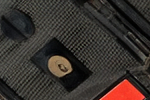} \hspace{\fsdurthree}  
                    \\ 
                    LQ \hspace{\fsdurthree} &
                    USRGAN \hspace{\fsdurthree} &
                    \makecell{DiffIR$_{S2}$ (Ours)} \hspace{\fsdurthree} 
                \end{tabular}
            \end{adjustbox}

            \begin{adjustbox}{valign=t}
                \Large
                \begin{tabular}{c}
                    \includegraphics[height=0.44\textwidth]{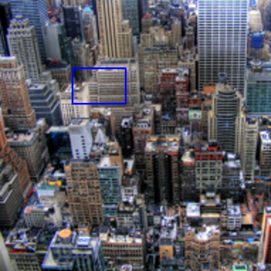} 
                \end{tabular}
                
            \end{adjustbox}
            
            \begin{adjustbox}{valign=t}
                \begin{tabular}{cccc}
                    \includegraphics[width=\widthscalefive \textwidth]{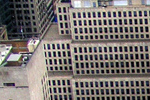} \hspace{\fsdurthree} &
                    \includegraphics[width=\widthscalefive \textwidth]{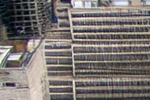} \hspace{\fsdurthree} &
                    \includegraphics[width=\widthscalefive \textwidth]{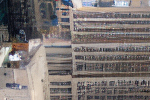} \hspace{\fsdurthree} 
                    \\
                    HQ \hspace{\fsdurthree} &
                    \makecell{BebyGAN} \hspace{\fsdurthree} &
                    \makecell{LDM} \hspace{\fsdurthree} 
                    \\
                    \includegraphics[width=\widthscalefive \textwidth]{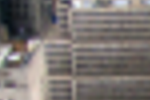} 
                    \hspace{\fsdurthree} &
                    \includegraphics[width=\widthscalefive \textwidth]{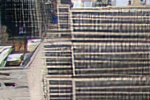} \hspace{\fsdurthree} &
                    \includegraphics[width=\widthscalefive \textwidth]{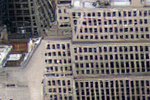} \hspace{\fsdurthree}  
                    \\ 
                    LQ \hspace{\fsdurthree} &
                    USRGAN \hspace{\fsdurthree} &
                    \makecell{DiffIR$_{S2}$ (Ours)} \hspace{\fsdurthree} 
                \end{tabular}
            \end{adjustbox}
            \\
            \begin{adjustbox}{valign=t}
                \Large
                \begin{tabular}{c}
                    \includegraphics[height=0.44\textwidth]{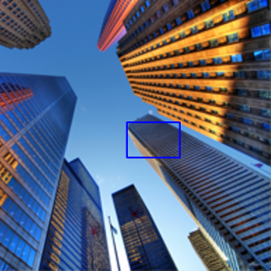} 
                \end{tabular}
                
            \end{adjustbox}
            
            \begin{adjustbox}{valign=t}
                \begin{tabular}{cccc}
                    \includegraphics[width=\widthscalefive \textwidth]{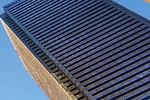} \hspace{\fsdurthree} &
                    \includegraphics[width=\widthscalefive \textwidth]{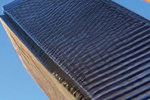} \hspace{\fsdurthree} &
                    \includegraphics[width=\widthscalefive \textwidth]{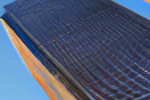} \hspace{\fsdurthree} 
                    \\
                    HQ \hspace{\fsdurthree} &
                    \makecell{BebyGAN} \hspace{\fsdurthree} &
                    \makecell{LDM} \hspace{\fsdurthree} 
                    \\
                    \includegraphics[width=\widthscalefive \textwidth]{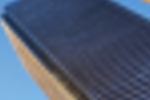} 
                    \hspace{\fsdurthree} &
                    \includegraphics[width=\widthscalefive \textwidth]{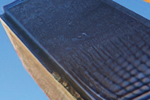} \hspace{\fsdurthree} &
                    \includegraphics[width=\widthscalefive \textwidth]{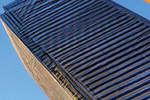} \hspace{\fsdurthree}  
                    \\ 
                    LQ \hspace{\fsdurthree} &
                    USRGAN \hspace{\fsdurthree} &
                    \makecell{DiffIR$_{S2}$ (Ours)} \hspace{\fsdurthree} 
                \end{tabular}
            \end{adjustbox}

            \begin{adjustbox}{valign=t}
                \Large
                \begin{tabular}{c}
                    \includegraphics[height=0.44\textwidth]{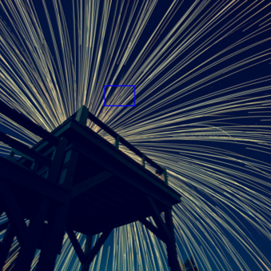} 
                \end{tabular}
                
            \end{adjustbox}
            
            \begin{adjustbox}{valign=t}
                \begin{tabular}{cccc}
                    \includegraphics[width=\widthscalefive \textwidth]{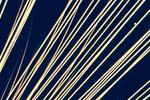} \hspace{\fsdurthree} &
                    \includegraphics[width=\widthscalefive \textwidth]{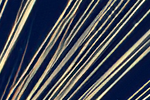} \hspace{\fsdurthree} &
                    \includegraphics[width=\widthscalefive \textwidth]{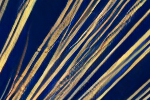} \hspace{\fsdurthree} 
                    \\
                    HQ \hspace{\fsdurthree} &
                    \makecell{BebyGAN} \hspace{\fsdurthree} &
                    \makecell{LDM} \hspace{\fsdurthree} 
                    \\
                    \includegraphics[width=\widthscalefive \textwidth]{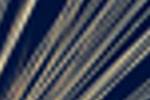} 
                    \hspace{\fsdurthree} &
                    \includegraphics[width=\widthscalefive \textwidth]{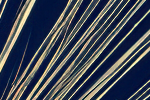} \hspace{\fsdurthree} &
                    \includegraphics[width=\widthscalefive \textwidth]{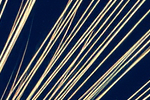} \hspace{\fsdurthree}  
                    \\ 
                    LQ \hspace{\fsdurthree} &
                    USRGAN \hspace{\fsdurthree} &
                    \makecell{DiffIR$_{S2}$ (Ours)} \hspace{\fsdurthree} 
                \end{tabular}
            \end{adjustbox}
        \end{tabular}
    }
    \caption{ Visual comparison of 4$\times$ \textbf{image super-resolution} methods. Zoom-in for better details.}
    \label{fig:SR_show}
    \vspace{-2mm}
\end{figure*}

\subsection{Evaluation on Inpainting}
We train and validate our DiffIR$_{S2}$ on inpainting using the same settings of LaMa~\cite{LaMa}. Specifically, we train our DiffIR with the batch size of 30 and patch size of 256 on Places-Standard~\cite{places2} and CelebA-HQ~\cite{celeba} datasets, respectively. We compare our DiffIR$_{S2}$ with SOTA inpainting methods (ICT~\cite{ICT}, LaMa~\cite{LaMa}, and RePaint~\cite{repaint}) using LPIPS~\cite{LPIPS} and FID~\cite{FID} on validation datasets.  

The quantitative results are shown in Tab.~\ref{tab:inpainting} and Fig.~\ref{fig:efficiency}~(a). We can see that our DiffIR$_{S2}$ significantly outperforms other methods. Specifically,  our DiffIR$_{S2}$ surpasses competitive method LaMa by a FID
margin of up to 0.2706 and 0.5583 with wide masks on Places and CelebA-HQ consuming similar total numbers of parameters and Mult-Adds. Furthermore, 
compared with DM based method RePaint~\cite{LDM}, our DiffIR$_{S2}$ can achieve better performance while merely consuming $4.3\%$ parameters and $0.1\%$ computational resources. This indicates that  DiffIR can fully and efficiently use the data estimation ability of DM for IR.

The qualitative results are shown in Fig.~\ref{fig:inpainting_show}. Our DiffIR$_{S2}$ can produce more realistic and reasonable structures and details than other competitive inpainting methods. More qualitative results are provided in the supplementary material.

\subsection{Evaluation on Image Super-Resolution}
We train and validate our DiffIR$_{S2}$ on image super-resolution. Specifically, we train DiffIR$_{S2}$ on DIV2K~\cite{DIV2K} (800 images) and Flickr2K~\cite{Flickr2K} (2650 images) datasets for $4\times$ super-resolution. The batch sizes are set to 64, and the LQ patch sizes are 64$\times$64. We evaluate our DiffIR$_{S2}$ and other SOTA GAN-based SR methods on five benchmarks (Set5~\cite{Set5}, Set14~\cite{Set14}, General100~\cite{general100}, Urban100~\cite{Urban100}, and DIV2K100~\cite{DIV2K}) using LPIPS~\cite{LPIPS} and PSNR.

Tab.~\ref{tab:SR} and Fig.~\ref{fig:efficiency}~(b) show the performance and Mult-Adds comparsion of  DiffIR$_{S2}$ with SOTA GAN-based SR methods: SFTGAN~\cite{SFTGAN}, SRGAN~\cite{SRGAN}, ESRGAN~\cite{ESRGAN}, USRGAN~\cite{USRGAN}, SPSR~\cite{SPSR}, and BebyGAN~\cite{BebyGAN}. We can see that DiffIR$_{S2}$ achieves the best performance. Compared with the competitive SR method BebyGAN, our DiffIR$_{S2}$ surpasses it by LPIPS margin of up to 0.0151 and 0.0089 on DIV2K100 and Urban100 while merely consuming $63\%$ computational resources. Moreover, it is notable that DiffIR$_{S2}$ significantly outperforms DM-based method LDM while consuming $2\%$ computational resources.

The qualitative results are shown in Fig.~\ref{fig:SR_show}. DiffIR$_{S2}$ achieves the best visual quality containing more realistic details. These visual comparisons are consistent with the quantitative results, showing the superiority of DiffIR. DiffIR can efficiently use the powerful DM to restore images. More visual results are given in supplementary material.

\begin{table}[t]
  \centering
  \caption{Quantitative comparison for \textbf{Single image motion deblurring} on benchmark datasets. Best and second best performance are marked in bold and underlined, respectively.}
  \resizebox{1\linewidth}{!}{
    \begin{tabular}{l|cc|cc}
    \toprule[0.2em]
    \multirow{2}[2]{*}{\textbf{Method}} & \multicolumn{2}{c|}{\textbf{GoPro}~\cite{gopro}} & \multicolumn{2}{c}{\textbf{HIDE}~\cite{hide}} \\
          & \multicolumn{1}{c}{PSNR $\uparrow$} & \multicolumn{1}{c|}{SSIM $\uparrow$} & \multicolumn{1}{c}{PSNR $\uparrow$} & \multicolumn{1}{c}{SSIM $\uparrow$} \\
    \midrule[0.2em]
    Xu \etal~\cite{deblur-xu} & \multicolumn{1}{c}{21.00} & \multicolumn{1}{c|}{0.741} & \multicolumn{1}{c}{-} & \multicolumn{1}{c}{-} \\
    DeblurGAN~\cite{deblurgan} & \multicolumn{1}{c}{28.70} & \multicolumn{1}{c|}{0.858} & \multicolumn{1}{c}{24.51} & \multicolumn{1}{c}{0.871} \\
    Nah \etal~\cite{gopro} & \multicolumn{1}{c}{29.08} & \multicolumn{1}{c|}{0.914} & \multicolumn{1}{c}{25.73} & \multicolumn{1}{c}{0.874} \\
    Zhang \etal~\cite{deblur-zhang} & \multicolumn{1}{c}{29.19} & \multicolumn{1}{c|}{0.931} & \multicolumn{1}{c}{-} & \multicolumn{1}{c}{-} \\
    DeblurGAN-v2~\cite{deblurganv2} & \multicolumn{1}{c}{29.55} & \multicolumn{1}{c|}{0.934} & \multicolumn{1}{c}{26.61} & \multicolumn{1}{c}{0.875} \\
    SRN~\cite{SRN}   & \multicolumn{1}{c}{30.26} & \multicolumn{1}{c|}{0.934} & \multicolumn{1}{c}{28.36} & \multicolumn{1}{c}{0.915} \\
    Gao \etal~\cite{deblur-gao} & \multicolumn{1}{c}{30.90} & \multicolumn{1}{c|}{0.935} & \multicolumn{1}{c}{29.11} & \multicolumn{1}{c}{0.913} \\
    DBGAN~\cite{DBGAN} & \multicolumn{1}{c}{31.10} & \multicolumn{1}{c|}{0.942} & \multicolumn{1}{c}{28.94} & \multicolumn{1}{c}{0.915} \\
    MT-RNN~\cite{MT-RNN} & \multicolumn{1}{c}{31.15} & \multicolumn{1}{c|}{0.945} & \multicolumn{1}{c}{29.15} & \multicolumn{1}{c}{0.918} \\
    DMPHN~\cite{DMPHN} & \multicolumn{1}{c}{31.20} & \multicolumn{1}{c|}{0.940} & \multicolumn{1}{c}{29.09} & \multicolumn{1}{c}{0.924} \\
    Suin \etal~\cite{deblur-suin} & \multicolumn{1}{c}{31.85} & \multicolumn{1}{c|}{0.948} & \multicolumn{1}{c}{29.98} & \multicolumn{1}{c}{0.930} \\
    MIMO-Unet+~\cite{MIMO-Unet} & \multicolumn{1}{c}{32.45} & \multicolumn{1}{c|}{0.957} & \multicolumn{1}{c}{29.99} & \multicolumn{1}{c}{0.930} \\
    IPT~\cite{IPT}   & \multicolumn{1}{c}{32.52} & \multicolumn{1}{c|}{-} & \multicolumn{1}{c}{-} & \multicolumn{1}{c}{-} \\
    MPRNet~\cite{MPRNet} & \multicolumn{1}{c}{32.66} & \multicolumn{1}{c|}{0.959} & \multicolumn{1}{c}{30.96} & \multicolumn{1}{c}{0.939} \\
    Restormer~\cite{restormer} & \multicolumn{1}{c}{\underline{32.92}} & \multicolumn{1}{c|}{\underline{0.961}} & \multicolumn{1}{c}{\underline{31.22}} & \multicolumn{1}{c}{\underline{0.942}} \\
    \midrule[0.2em]
    DiffIR$_{S2}$ (Ours) &   \multicolumn{1}{c}{\textbf{33.20}}    &       \multicolumn{1}{c|}{\textbf{0.963}}    & \multicolumn{1}{c}{\textbf{31.55}} & \multicolumn{1}{c}{\textbf{0.947}} \\
    \bottomrule[0.2em]
    \end{tabular}%
    }
  \label{tab:deblur}%
  \vspace{-4mm}
\end{table}%

\begin{figure*}[t]
    \setlength{\fsdurthree}{0mm}
    \LARGE
    \centering
   \resizebox{1\linewidth}{!}{
        \begin{tabular}{cc}
            \begin{adjustbox}{valign=t}
                \Large
                \begin{tabular}{c}
                    \includegraphics[height=0.44\textwidth]{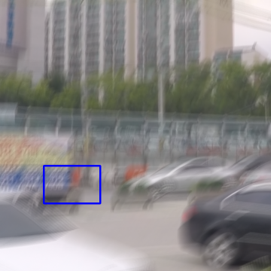} 
                \end{tabular}
                
            \end{adjustbox}
            
            \begin{adjustbox}{valign=t}
                \begin{tabular}{cccc}
                    \includegraphics[width=\widthscalefive \textwidth]{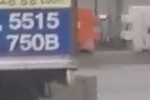} \hspace{\fsdurthree} &
                    \includegraphics[width=\widthscalefive \textwidth]{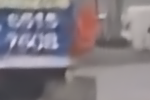} \hspace{\fsdurthree} &
                    \includegraphics[width=\widthscalefive \textwidth]{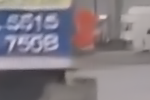} \hspace{\fsdurthree} 
                    \\
                    HQ \hspace{\fsdurthree} &
                    \makecell{MT-RNN} \hspace{\fsdurthree} &
                    \makecell{Restormer} \hspace{\fsdurthree} 
                    \\
                    \includegraphics[width=\widthscalefive \textwidth]{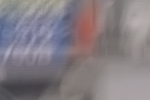} 
                    \hspace{\fsdurthree} &
                    \includegraphics[width=\widthscalefive \textwidth]{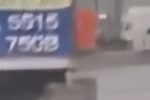} \hspace{\fsdurthree} &
                    \includegraphics[width=\widthscalefive \textwidth]{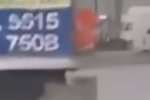} \hspace{\fsdurthree}  
                    \\ 
                    LQ \hspace{\fsdurthree} &
                    MPRNet \hspace{\fsdurthree} &
                    \makecell{DiffIR$_{S2}$ (Ours)} \hspace{\fsdurthree} 
                \end{tabular}
            \end{adjustbox}

            \begin{adjustbox}{valign=t}
                \Large
                \begin{tabular}{c}
                    \includegraphics[height=0.44\textwidth]{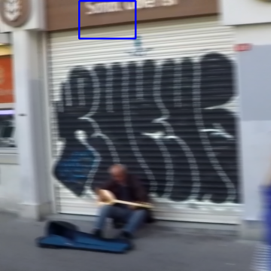} 
                \end{tabular}
                
            \end{adjustbox}
            
            \begin{adjustbox}{valign=t}
                \begin{tabular}{cccc}
                    \includegraphics[width=\widthscalefive \textwidth]{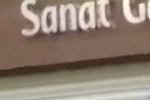} \hspace{\fsdurthree} &
                    \includegraphics[width=\widthscalefive \textwidth]{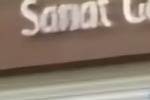} \hspace{\fsdurthree} &
                    \includegraphics[width=\widthscalefive \textwidth]{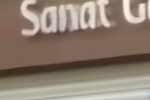} \hspace{\fsdurthree} 
                    \\
                    HQ \hspace{\fsdurthree} &
                    \makecell{MT-RNN} \hspace{\fsdurthree} &
                    \makecell{Restormer} \hspace{\fsdurthree} 
                    \\
                    \includegraphics[width=\widthscalefive \textwidth]{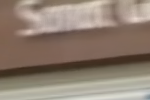} 
                    \hspace{\fsdurthree} &
                    \includegraphics[width=\widthscalefive \textwidth]{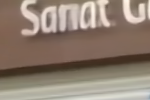} \hspace{\fsdurthree} &
                    \includegraphics[width=\widthscalefive \textwidth]{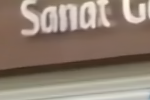} \hspace{\fsdurthree}  
                    \\ 
                    LQ \hspace{\fsdurthree} &
                    MPRNet \hspace{\fsdurthree} &
                    \makecell{DiffIR$_{S2}$ (Ours)} \hspace{\fsdurthree} 
                \end{tabular}
            \end{adjustbox}
            \\
            \begin{adjustbox}{valign=t}
                \Large
                \begin{tabular}{c}
                    \includegraphics[height=0.44\textwidth]{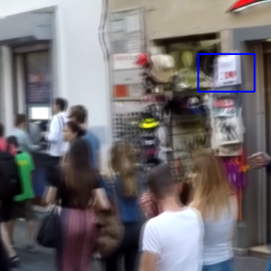} 
                \end{tabular}
                
            \end{adjustbox}
            
            \begin{adjustbox}{valign=t}
                \begin{tabular}{cccc}
                    \includegraphics[width=\widthscalefive \textwidth]{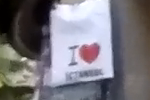} \hspace{\fsdurthree} &
                    \includegraphics[width=\widthscalefive \textwidth]{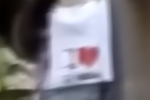} \hspace{\fsdurthree} &
                    \includegraphics[width=\widthscalefive \textwidth]{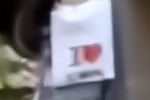} \hspace{\fsdurthree} 
                    \\
                    HQ \hspace{\fsdurthree} &
                    \makecell{MT-RNN} \hspace{\fsdurthree} &
                    \makecell{Restormer} \hspace{\fsdurthree} 
                    \\
                    \includegraphics[width=\widthscalefive \textwidth]{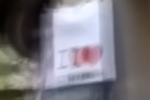} 
                    \hspace{\fsdurthree} &
                    \includegraphics[width=\widthscalefive \textwidth]{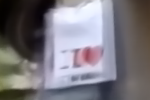} \hspace{\fsdurthree} &
                    \includegraphics[width=\widthscalefive \textwidth]{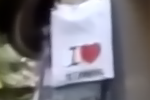} \hspace{\fsdurthree}  
                    \\ 
                    LQ \hspace{\fsdurthree} &
                    MPRNet \hspace{\fsdurthree} &
                    \makecell{DiffIR$_{S2}$ (Ours)} \hspace{\fsdurthree} 
                \end{tabular}
            \end{adjustbox}

            \begin{adjustbox}{valign=t}
                \Large
                \begin{tabular}{c}
                    \includegraphics[height=0.44\textwidth]{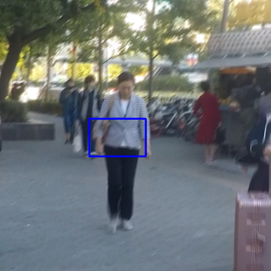} 
                \end{tabular}
                
            \end{adjustbox}
            
            \begin{adjustbox}{valign=t}
                \begin{tabular}{cccc}
                    \includegraphics[width=\widthscalefive \textwidth]{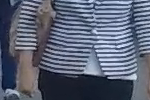} \hspace{\fsdurthree} &
                    \includegraphics[width=\widthscalefive \textwidth]{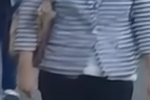} \hspace{\fsdurthree} &
                    \includegraphics[width=\widthscalefive \textwidth]{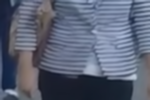} \hspace{\fsdurthree} 
                    \\
                    HQ \hspace{\fsdurthree} &
                    \makecell{MT-RNN} \hspace{\fsdurthree} &
                    \makecell{Restormer} \hspace{\fsdurthree} 
                    \\
                    \includegraphics[width=\widthscalefive \textwidth]{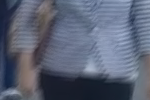} 
                    \hspace{\fsdurthree} &
                    \includegraphics[width=\widthscalefive \textwidth]{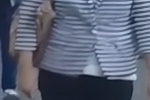} \hspace{\fsdurthree} &
                    \includegraphics[width=\widthscalefive \textwidth]{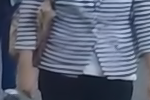} \hspace{\fsdurthree}  
                    \\ 
                    LQ \hspace{\fsdurthree} &
                    MPRNet \hspace{\fsdurthree} &
                    \makecell{DiffIR$_{S2}$ (Ours)} \hspace{\fsdurthree} 
                \end{tabular}
            \end{adjustbox}
        \end{tabular}
    }
    \caption{Visual comparison of \textbf{single image motion deblurring} methods. Zoom-in for better details.}
    \label{fig:deblur_show}
    \vspace{-3mm}
\end{figure*}

\subsection{Evaluation on Image Motion Deblurring}
We train DiffIR on GoPro~\cite{gopro} dataset for image motion deblurring and evaluate DiffIR  on two classic benchmarks (GoPro, HIDE~\cite{hide}). We compare DiffIR$_{S2}$ with the state-of-the-art image motion deblurring methods, including Restormer~\cite{restormer}, MPRNet~\cite{MPRNet}, and IPT~\cite{IPT}.  

The quantitative results (PSNR and SSIM) are shown in Tab.~\ref{tab:deblur}, and the Mult-Adds are shown in Fig.~\ref{fig:efficiency}~(c). We can see that our DiffIR$_{S2}$ outperforms other motion deblurring methods. Specifically, DiffIR$_{S2}$  surpasses IPT and MIMI-Unet+ by 0.68~dB and 0.54~dB on GoPro, respectively. Furthermore, DiffIR$_{S2}$ surpasses Restormer by 0.28~dB and 0.33~dB on GoPro and HIDE datasets separately, only consuming $78\%$ computational resources. This demonstrates the effectiveness of DiffIR.

The qualitative results are shown in Fig.~\ref{fig:deblur_show}, and our DiffIR$_{S2}$ has the best visual quality containing more realistic details close to corresponding HQ images. More qualitative results are provided in the supplementary material.

\vspace{-1mm}
\section{Ablation Study}
\label{sec:ablation}
\noindent\textbf{Efficient diffusion model for image restoration.}
In this part, we validate the effectiveness of the components in DiffIR, such as DM, training schemes for DM, and whether inserting variance noise in DM (Tab.~\ref{tab:DiffIR}).  

\textbf{(1)} DiffIR$_{S2}$-V3 is actually the DiffIR$_{S2}$ adopted in Tab.~\ref{tab:inpainting}, and DiffIR$_{S1}$ is the first stage pretraining network with ground-truth images as inputs. Comparing DiffIR$_{S1}$ and DiffIR$_{S2}$-V3, we can see that  DiffIR$_{S2}$-V3 has quite similar LPIPS with DiffIR$_{S1}$, which means that DM has powerful data modeling ability to predict accurate IPR. 

\textbf{(2)} To further demonstrate the effectiveness of DM, we cancel using DM in DiffIR$_{S2}$-V3 to obtain DiffIR$_{S2}$-V1. Comparing DiffIR$_{S2}$-V1 and DiffIR$_{S2}$-V3, we can see that DiffIR$_{S2}$-V3 (using DM) significantly outperform DiffIR$_{S2}$-V1. That means the IPR learned by DM can effectively guide DIRformer to restore LQ images.

\textbf{(3)} To explore the better training schemes for DM, we compare two training schemes: traditional DM optimization and our proposed joint optimization. Since traditional DM~\cite{LDM,DDPM} requires many iterations to estimate large images or feature maps, they have to adopt traditional DM optimization by randomly sampling a timestep to optimize the denoising network, which cannot optimize with the later decoder (\ie, DIRformer in our paper). Since DiffIR merely uses DM to estimate a compact one-dimensional vector IPR, we can use several times iterations to obtain quite accurate results. Therefore, we can adopt joint optimization by running all iterations of the denoising network to obtain IPR to optimize with DIRformer jointly. Comparing DiffIR$_{S2}$-V2 and DiffIR$_{S2}$-V3, DiffIR$_{S2}$-V3 significantly surpass the DiffIR$_{S2}$-V2, which demonstrates the effectiveness of our proposed joint optimization for training DM. That is because the DM's minor estimation error in IPR  may lead to the performance drop of the DIRformer. Training DM and DIRformer jointly can address this problem.

\textbf{(4)} In traditional DM methods, they will insert variance noise in the reverse DM process (Eq.~\eqref{eq:diff3}) to generate more realistic images. Different from traditional DM predicting images or feature maps, we use DM to estimate IPR.  In DiffIR$_{S2}$-V4, we insert noise in the reverse DM process. As we can see, DiffIR$_{S2}$-V3 achieve better performance than DiffIR$_{S2}$-V4. That means it is better to cancel inserting noise to guarantee the accuracy of the estimated IPR.

\begin{table*}[htbp]
  \centering
  \caption{FID results evaluated on CelebA-HQ for inpainting. The performance and Mult-Adds are measured on an LQ size of 256$\times$256.}
  \resizebox{1\linewidth}{!}{
    \begin{tabular}{l|c|ccccc|c}
    \toprule[0.2em]
    \multirow{2}[4]{*}{\textbf{Method}} & \multirow{2}[4]{*}{\textbf{\shortstack{Mult-Adds (G)}}} & \multirow{2}[4]{*}{\textbf{GT}} & \multirow{2}[4]{*}{\textbf{DM}} & \multicolumn{2}{c}{\textbf{Training Schemes}} & \multirow{2}[4]{*}{\textbf{\shortstack{Inserting\\ Noise}} } & \multirow{2}[4]{*}{\textbf{CelebA-HQ} } \\
\cmidrule{5-6}          &       &       &       & \textbf{\shortstack{Traditional DM \\Optimization}}  & \textbf{\shortstack{Joint \\Optimization}} &       &  \\
    \midrule
    DiffIR$_{S1}$ &   47.97    & \Checkmark     & \XSolidBrush     & \XSolidBrush     & \XSolidBrush     & \XSolidBrush     & 4.8045 \\
    \midrule
    DiffIR$_{S2}$-V1 &   51.63    & \XSolidBrush     & \XSolidBrush     & \XSolidBrush     & \XSolidBrush     & \XSolidBrush     &  5.6782\\
    DiffIR$_{S2}$-V2 &    51.63   & \XSolidBrush     & \Checkmark     & \Checkmark     & \XSolidBrush     & \XSolidBrush     &  5.9766\\
    DiffIR$_{S2}$-V3 (Ours) &    51.63   & \XSolidBrush     & \Checkmark     & \XSolidBrush     & \Checkmark     & \XSolidBrush     &  5.1440\\
    DiffIR$_{S2}$-V4 &   51.63    & \XSolidBrush     & \Checkmark     & \XSolidBrush     & \Checkmark     & \Checkmark     & 5.1937 \\
    \bottomrule[0.2em]
    \end{tabular}%
}
  \label{tab:DiffIR}%
  \vspace{-4mm}
\end{table*}%

\begin{table}[t]
  \centering
  \caption{ DM loss functions comparison (FID) in inpainting.}
  \resizebox{1\linewidth}{!}{
    \begin{tabular}{l|ccc}
    \toprule[0.2em]
    \textbf{Loss}  & $\mathcal{L}_{diff}$ (Eq.~\eqref{eq:diff})   & $\mathcal{L}_{2}$ (Eq.~\eqref{eq:l2})    & $\mathcal{L}_{kl}$ (Eq.~\eqref{eq:kl}) \\
    \midrule
    \textbf{CelebA-HQ}$\downarrow$ & 5.1440     & 5.1837     & 5.2365 \\
    \bottomrule[0.2em]
    \end{tabular}%
    }
  \label{tab:loss}%
  \vspace{-4mm}
\end{table}%

\noindent\textbf{The loss functions for DM.}
We explore which loss function is best to guide the denoising network and CPEN$_{S2}$ to learn to estimate accurate IPR from LQ images. Here, we define three loss functions. \textbf{(1)} We define $\mathcal{L}_{diff}$ for optimization (Eq.~\eqref{eq:diff}). \textbf{(2)} We adopt $\mathcal{L}_{2}$  (Eq.~\eqref{eq:l2}) to measure estimation error.  \textbf{(3)} We use the Kullback Leibler divergence to measure distribution similarity ($\mathcal{L}_{kl}$, Eq.~\eqref{eq:kl}).
\vspace{-2mm}
\begin{equation}
\label{eq:l2}
\vspace{-2mm}
\mathcal{L}_{2}=\frac{1}{4C^{\prime}}\sum_{i=1}^{4C^{\prime}}\left(\hat{\mathbf{Z}}(i)-\mathbf{Z}(i)\right)^{2}, 
\end{equation}
\begin{equation}
\label{eq:kl}
\vspace{-1mm}
\begin{aligned}
\mathcal{L}_{kl} &= \sum_{i=1}^{4C^{\prime}} \mathbf{Z}_{norm}(i) \log \left(\frac{\mathbf{Z}_{norm}(i)}{\hat{\mathbf{Z}}_{norm}(i)}\right),
\end{aligned}
\end{equation}

where $\hat{\mathbf{Z}}$ and $\mathbf{Z} \in \mathbb{R}^{4C^{\prime}}$ are IPRs extracted by DiffIR$_{S1}$ and DiffIR$_{S2}$ respectively. $\hat{\mathbf{Z}}_{norm}$ and $\mathbf{Z}_{norm}\in \mathbb{R}^{4C^{\prime}}$ are normalized with softmax operation of $\hat{\mathbf{Z}}$ and $\mathbf{Z}$ separately. We apply these three loss functions on DiffIR$_{S2}$ separately to learn to directly estimate the accurate IPR from LQ images. Then, we evaluate them on CelebA-HQ in the inpainting task. The results are shown in Tab.~\ref{tab:loss}. We can see that the performance of $\mathcal{L}_{diff}$ is better than $\mathcal{L}_{2}$ and $\mathcal{L}_{kl}$.    

\begin{figure}[t]
	\centering
	\includegraphics[height=4cm]{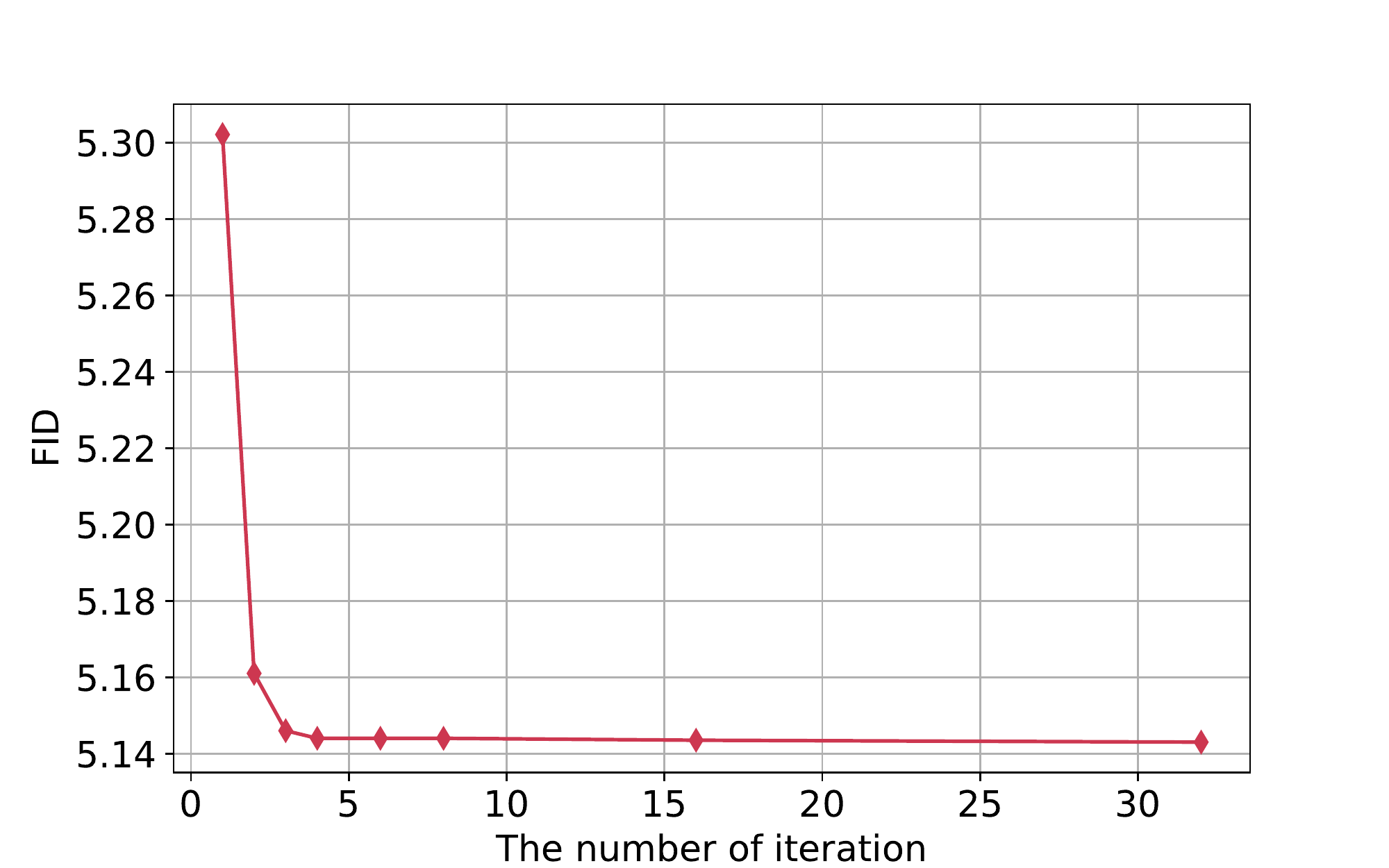}
	\caption{Ablation study of the number of iterations in DM.  }
	\label{fig:iter}
 \vspace{-4mm}
\end{figure}

\noindent\textbf{Impact of the number of iterations.}
In this part, we explore how the number of iterations in DM affects the performance of DiffIR$_{S2}$. We set different number of iterations in DiffIR$_{S2}$ and tune the $\beta_{t}$ ($\alpha_t=1-\beta_{t}$) in Eq.~\eqref{eq:mydiff1} to make $\mathbf{Z}$ be  Gaussian noise $\mathbf{Z}_{T} \sim \mathcal{N}(0,1)$ after diffusion process (\ie, $\bar{\alpha}_{T} \rightarrow0$). The results are shown in Fig.~\ref{fig:iter}. As iterations increase to 3, the performance of DiffIR$_{S2}$ will significantly improve. As the number of iteration is larger than 4, DiffIR$_{S2}$ almost keep stable, which means it reaches the upper bound. Besides, we can see that our DiffIR$_{S2}$ has more quick convergence speed than traditional DM (requiring more than 200 iterations). That is because we merely perform DM on IPR (a compact one-dimensional vector).     

\vspace{-1mm}
\section{Conclusion}
Traditional DMs achieve impressive performance in image synthesis. Different from image synthesis generating each pixel from scratch, IR gives an LQ image as a reference. Thus, it is inefficient to directly apply the traditional DM paradigm to IR. In this paper, we propose an efficient diffusion model for IR (\ie, DiffIR), consisting of CPEN, DIRformer, and denoising network. Specifically, we first input ground-truth image into CPEN$_{S1}$ to generate a compact IPR to guide DIRformer. After that, we train DM to estimate the IPR extracted by CPEN$_{S1}$. Compared with traditional DMs, our DiffIR can use much fewer iterations than traditional DMs to obtain accurate estimations and reduce artifacts in restored images. Furthermore, thanks to the few iterations, our DiffIR can adopt joint optimization of CPEN$_{S2}$, DIRformer, and denoising network to reduce the influence of estimation error. Extensive experiments show that DiffIR can achieve a general SOTA IR performance.

{\small
\balance
\bibliographystyle{ieee_fullname}
\bibliography{egbib}
}

\newpage
\appendix
\section{Appendix}

\section{ Evaluation on Real-world SR}
We train and validate our DiffIR$_{S2}$ on real-world SR using
the same settings of Real-ESRGAN~\cite{Real-ESRGAN}. Specifically, we adopt the same loss functions of Real-ESRGAN~\cite{ESRGAN}, which further introduce perceptual loss and adversarial loss to the basic $\mathcal{L}_{1}$ loss. We set the learning rate of the  DiffIR$_{S2}$ to $2\times10^{-4}$.   We further validate the effectiveness of  DiffIR$_{S2}$ on Real-World datasets. For optimization, we use Adam with $\beta_{1}=0.9$, $\beta_{2}=0.99$. In both two stages of training, we set the batch size to $64$, with the input patch size being $64$. We evaluate all methods on the dataset provided in the challenge of Real-World Super-Resolution:  NTIRE2020 Track1 and Tracks~\cite{NTIRE2020}. In addition, we also validate our DiffIR on RealSRSet~\cite{RealSR}. Since NTIRE2020 Track1  and RealSRSet datasets provide a paired validation set, we use the LPIPS~\cite{LPIPS}, DISTS~\cite{DISTS}, and PSNR for the evaluation. 

The quantitative results are shown in Tab.~\ref{tab:real}. We can see that  DiffIR$_{S2}$ outperforms SOTA real-world SR method KDSR$_{S}$-GAN on LPIPS, DISTS, and PSNR,  consuming fewer computational costs. In addition, we can see that  DiffIR$_{S2}$ outperforms classic real-world SR method Real-ESRGAN on LPIPS, DISTS, and PSNR, only consuming its $63\%$ Mult-Adds. Furthermore, compared with DM-based LDM~\cite{LDM}, DiffIR$_{S2}$ achieve much better performance consuming only $2\%$ Mult-Adds.

\begin{figure*}[t]
 \setlength{\fsdurthree}{0mm}
    \LARGE
    \centering
   \resizebox{0.8\linewidth}{!}{
        \begin{tabular}{cc}
            \begin{adjustbox}{valign=t}
                \Large
                \begin{tabular}{c}
                    \includegraphics[height=0.623\textwidth]{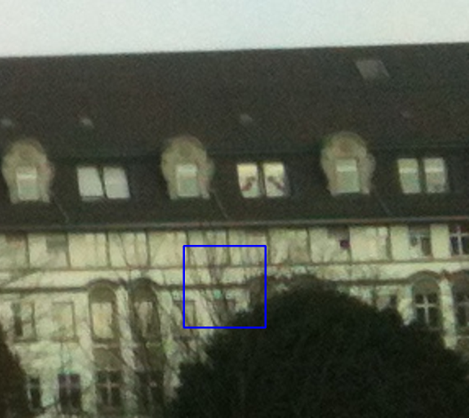} 
                \end{tabular}
                
            \end{adjustbox}
            
            \begin{adjustbox}{valign=t}
                \begin{tabular}{cccc}
                    \includegraphics[width=\widthscalefive \textwidth]{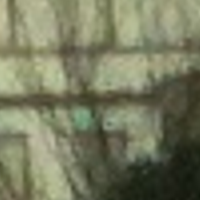} \hspace{\fsdurthree} &
                    \includegraphics[width=\widthscalefive \textwidth]{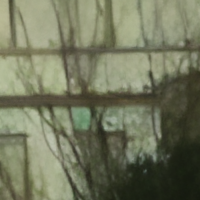} \hspace{\fsdurthree} &
                    \includegraphics[width=\widthscalefive \textwidth]{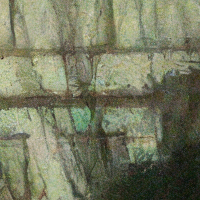} \hspace{\fsdurthree} 
                    \\
                    LQ \hspace{\fsdurthree} &
                    \makecell{KDSR$_{S}$-GAN~\cite{KDSR}} \hspace{\fsdurthree} &
                    \makecell{LDM~\cite{LDM}} \hspace{\fsdurthree} 
                    \\
                    \includegraphics[width=\widthscalefive \textwidth]{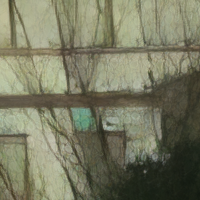} 
                    \hspace{\fsdurthree} &
                    \includegraphics[width=\widthscalefive \textwidth]{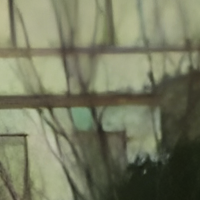} \hspace{\fsdurthree} &
                    \includegraphics[width=\widthscalefive \textwidth]{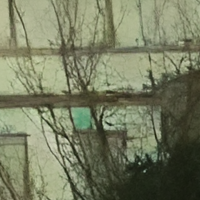} \hspace{\fsdurthree}  
                    \\ 
                    BSRGAN~\cite{BSRGAN} \hspace{\fsdurthree} &
                    Real-ESRGAN~\cite{Real-ESRGAN} \hspace{\fsdurthree} &
                    \makecell{DiffIR$_{S2}$ (Ours)} \hspace{\fsdurthree} 
                \end{tabular}
            \end{adjustbox}
            \\
            \begin{adjustbox}{valign=t}
                \Large
                \begin{tabular}{c}
                    \includegraphics[height=0.623\textwidth]{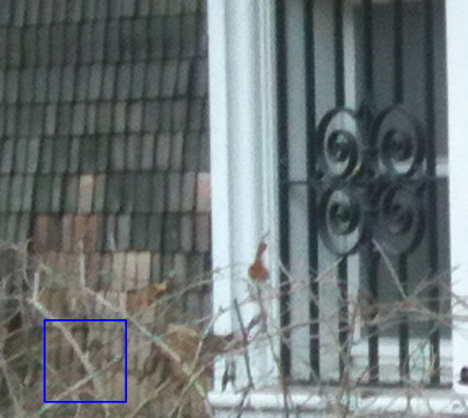} 
                \end{tabular}
                
            \end{adjustbox}
            
            \begin{adjustbox}{valign=t}
                \begin{tabular}{cccc}
                    \includegraphics[width=\widthscalefive \textwidth]{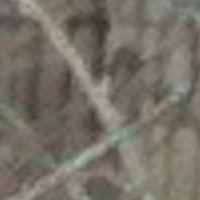} \hspace{\fsdurthree} &
                    \includegraphics[width=\widthscalefive \textwidth]{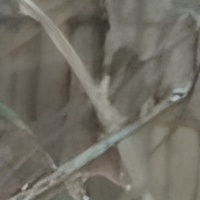} \hspace{\fsdurthree} &
                    \includegraphics[width=\widthscalefive \textwidth]{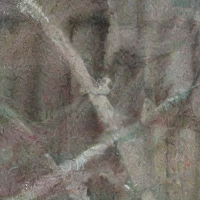} \hspace{\fsdurthree} 
                    \\
                    LQ \hspace{\fsdurthree} &
                    \makecell{KDSR$_{S}$-GAN~\cite{KDSR}} \hspace{\fsdurthree} &
                    \makecell{LDM~\cite{LDM}} \hspace{\fsdurthree} 
                    \\
                    \includegraphics[width=\widthscalefive \textwidth]{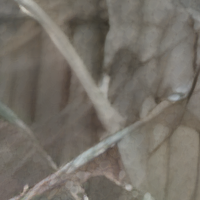} 
                    \hspace{\fsdurthree} &
                    \includegraphics[width=\widthscalefive \textwidth]{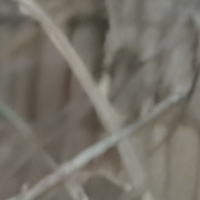} \hspace{\fsdurthree} &
                    \includegraphics[width=\widthscalefive \textwidth]{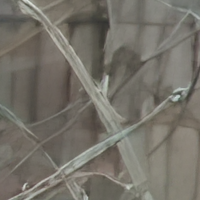} \hspace{\fsdurthree}  
                    \\ 
                    BSRGAN~\cite{BSRGAN} \hspace{\fsdurthree} &
                    Real-ESRGAN~\cite{Real-ESRGAN} \hspace{\fsdurthree} &
                    \makecell{DiffIR$_{S2}$ (Ours)} \hspace{\fsdurthree} 
                \end{tabular}
            \end{adjustbox}
            \\
            \begin{adjustbox}{valign=t}
                \Large
                \begin{tabular}{c}
                    \includegraphics[height=0.623\textwidth]{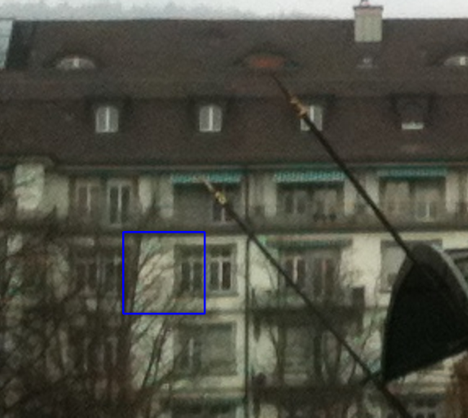} 
                \end{tabular}
                
            \end{adjustbox}
            
            \begin{adjustbox}{valign=t}
                \begin{tabular}{cccc}
                    \includegraphics[width=\widthscalefive \textwidth]{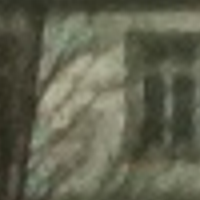} \hspace{\fsdurthree} &
                    \includegraphics[width=\widthscalefive \textwidth]{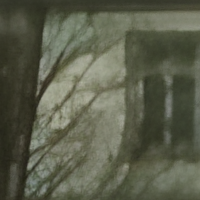} \hspace{\fsdurthree} &
                    \includegraphics[width=\widthscalefive \textwidth]{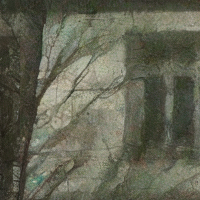} \hspace{\fsdurthree} 
                    \\
                    LQ \hspace{\fsdurthree} &
                    \makecell{KDSR$_{S}$-GAN~\cite{KDSR}} \hspace{\fsdurthree} &
                    \makecell{LDM~\cite{LDM}} \hspace{\fsdurthree} 
                    \\
                    \includegraphics[width=\widthscalefive \textwidth]{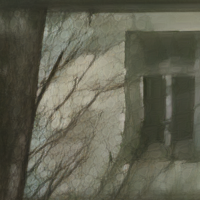} 
                    \hspace{\fsdurthree} &
                    \includegraphics[width=\widthscalefive \textwidth]{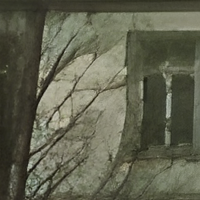} \hspace{\fsdurthree} &
                    \includegraphics[width=\widthscalefive \textwidth]{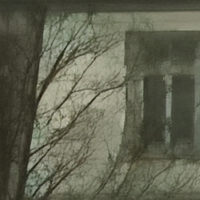} \hspace{\fsdurthree}  
                    \\ 
                    BSRGAN~\cite{BSRGAN} \hspace{\fsdurthree} &
                    Real-ESRGAN~\cite{Real-ESRGAN} \hspace{\fsdurthree} &
                    \makecell{DiffIR$_{S2}$ (Ours)} \hspace{\fsdurthree} 
                \end{tabular}
            \end{adjustbox}
            \\
            \begin{adjustbox}{valign=t}
                \Large
                \begin{tabular}{c}
                    \includegraphics[height=0.623\textwidth]{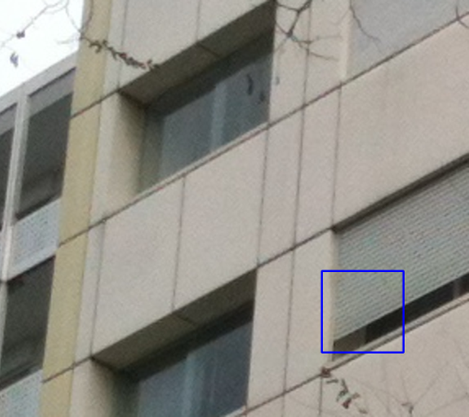} 
                \end{tabular}
                
            \end{adjustbox}
            
            \begin{adjustbox}{valign=t}
                \begin{tabular}{cccc}
                    \includegraphics[width=\widthscalefive \textwidth]{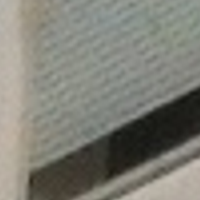} \hspace{\fsdurthree} &
                    \includegraphics[width=\widthscalefive \textwidth]{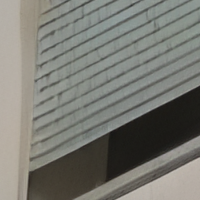} \hspace{\fsdurthree} &
                    \includegraphics[width=\widthscalefive \textwidth]{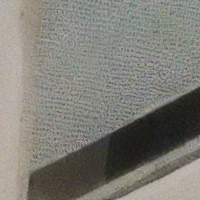} \hspace{\fsdurthree} 
                    \\
                    LQ \hspace{\fsdurthree} &
                    \makecell{KDSR$_{S}$-GAN~\cite{KDSR}} \hspace{\fsdurthree} &
                    \makecell{LDM~\cite{LDM}} \hspace{\fsdurthree} 
                    \\
                    \includegraphics[width=\widthscalefive \textwidth]{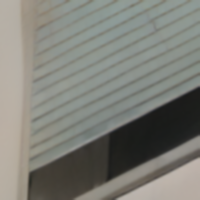} 
                    \hspace{\fsdurthree} &
                    \includegraphics[width=\widthscalefive \textwidth]{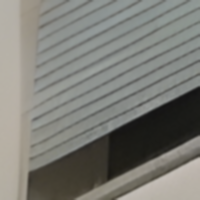} \hspace{\fsdurthree} &
                    \includegraphics[width=\widthscalefive \textwidth]{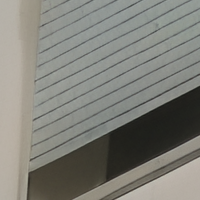} \hspace{\fsdurthree}  
                    \\ 
                    BSRGAN~\cite{BSRGAN} \hspace{\fsdurthree} &
                    Real-ESRGAN~\cite{Real-ESRGAN} \hspace{\fsdurthree} &
                    \makecell{DiffIR$_{S2}$ (Ours)} \hspace{\fsdurthree} 
                \end{tabular}
            \end{adjustbox}
        \end{tabular}
    }
    \caption{ Visual comparison of 4$\times$ \textbf{real-world super-resolution} methods. Zoom-in for better details.}
    \label{fig:sup_realSR_show}
\end{figure*}

We also visualize the results on NTIRE2020 Track2, which was captured with smartphones. The qualitative results are shown in Fig.~\ref{fig:sup_realSR_show}. We can see that DiffIR$_{S2}$ achieves the best performance.

\begin{table*}[t]
  \centering
  \caption{4$\times$ SR quantitative comparison on real-world SR benchmarks. The Mult-Adds are computed based on an
LR size of 256 $\times$ 256. Best and second best performance are marked in bold and underlined, respectively. The bottom two methods marked in gray adopt the diffusion model. }

    \begin{tabular}{l|c|ccc|ccc}
    \toprule[0.2em]
    \multirow{2}[2]{*}{Methods}  & \multirow{2}[2]{*}{Mult-Adds (T)}  & \multicolumn{3}{c|}{RealSRSet~\cite{RealSR}} & \multicolumn{3}{c}{NTIRE2020 Track1~\cite{NTIRE2020}} \\
          &                     & LPIPS$\downarrow$ & DISTS$\downarrow$ & PSNR$\uparrow$   & LPIPS$\downarrow$ & DISTS$\downarrow$ & PSNR$\uparrow$  \\
    \midrule
    BSRGAN~\cite{BSRGAN} & 1.18  & 0.3648 & 0.1676 & 26.90  & 0.3691 & 0.1368 & 26.75  \\
    Real-ESRGAN~\cite{Real-ESRGAN} & 1.18  & 0.3629 & \underline{0.1609} & 26.07  & 0.3471 & 0.1326 &26.40  \\
    KDSR$_{s}$-GAN~\cite{KDSR}  & 0.86  & \underline{0.3610} & 0.1627 &\underline{27.18}   & \underline{0.3198} & \underline{0.1252} & \underline{27.12} \\
    \rowcolor{lightgray}
    LDM~\cite{LDM}  & 37.25  & 0.4369 & 0.1982 & 26.37 & 0.4763 & 0.1844 & 25.68  \\
    \rowcolor{lightgray}
    DiffIR$_{S2}$ (Ours)  & 0.74  & \textbf{0.3527} & \textbf{0.1588} & \textbf{27.65}  & \textbf{0.3088} & \textbf{0.1131} & \textbf{27.31}  \\ 
    \bottomrule[0.2em]
    \end{tabular}%
    
  \label{tab:real}%
\end{table*}%

\vspace{-1mm}
\section{Algorithm}
\vspace{-1mm}
 The algorithm of DiffIR$_{2}$ training is summarized in Alg.~\ref{alg:train_DiffIR}. The algorithm of DiffIR$_{2}$ inference is summarized in Alg.~\ref{alg:infer_DiffIR}.

\begin{algorithm*}[t]
	\caption{ DiffIR$_{S2}$ Training}
	\label{alg:train_DiffIR}
	\textbf{Input}: Trained DiffIR$_{S1}$ (including CPEN$_{S1}$ and DIRformer), $\beta_t (t\in[1,T])$. \\
	\textbf{Output}: Trained DiffIR$_{S2}$. \\
        \vspace{-4mm}
	\begin{algorithmic}[1] 
		\STATE Init: $\alpha_t=1-\beta_t$, $\bar{\alpha}_T=\prod_{i=0}^T \alpha_i$.
            \STATE Init: The DIRformer of DiffIR$_{S2}$ copies the parameters of trained DiffIR$_{S1}$. 
            \FOR{$I_{LQ}$,  $I_{GT}$ }
            \STATE    $\mathbf{Z}=\operatorname{CPEN_{S1}}(\operatorname{PixelUnshuffle}(\operatorname{Concat}(I_{GT},I_{LQ}))). $ (paper Eq.~(\textbf{\textcolor{blue}{5}}))
            \STATE \textbf{Diffusion Process}:
            \STATE  We sample $\mathbf{Z}_{T}$ by $q\left(\mathbf{Z}_T \mid               \mathbf{Z}\right)=\mathcal{N}\left(\mathbf{Z}_T; \sqrt{\bar{\alpha}_T} \mathbf{Z},\left(1-\bar{\alpha}_T\right) \mathbf{I}\right)$  (\ie, diffusion process. paper Eq.~(\textbf{\textcolor{blue}{10}})) 
            \STATE \textbf{Reverse Process}:
            \STATE $\mathbf{\hat{Z}}_T = \mathbf{Z}_{T}$
            \STATE $\mathbf{D}=\operatorname{CPEN_{S2}}(\operatorname{PixelUnshuffle}(I_{LQ}))$ (paper Eq.~(\textbf{\textcolor{blue}{12}}))
            \FOR{$t=T$ to $1$ }
            \STATE  $\mathbf{\hat{Z}}_{t-1}=\frac{1}{\sqrt{\alpha_t}}\left(\mathbf{\hat{Z}}_t-\epsilon_{\theta}(\operatorname{Concat}(\mathbf{\hat{Z}}_t,t,\mathbf{D})) \frac{1-\alpha_t}{\sqrt{1-\bar{\alpha}_t}}\right)$ (paper Eq.~(\textbf{\textcolor{blue}{11}}))               
            \ENDFOR
            \STATE $\mathbf{\hat{Z}}=\mathbf{\hat{Z}}_{0}$
            \STATE $\hat{I}_{HQ} = \operatorname{DIRformer}(I_{LQ},\mathbf{\hat{Z}})$
            \STATE Calculate $\mathcal{L}_{diff}$ loss (paper Eq.~(\textbf{\textcolor{blue}{13}})).
            \ENDFOR     
		\STATE Output the trained model DiffIR$_{S2}$.
	\end{algorithmic}
\end{algorithm*}

\begin{algorithm*}[t]
	\caption{ DiffIR$_{S2}$ Inference}
	\label{alg:infer_DiffIR}
	\textbf{Input}: Trained DiffIR$_{S2}$ (including CPEN$_{S2}$ and DIRformer), $\beta_t (t\in[1,T])$, LQ images $I_{LQ}$. \\
	\textbf{Output}: Restored HQ images $\hat{I}_{HQ}$. \\
        \vspace{-4mm}
	\begin{algorithmic}[1] 
		\STATE Init: $\alpha_t=1-\beta_t$, $\bar{\alpha}_T=\prod_{i=0}^T \alpha_i$.
            \STATE \textbf{Reverse Process}:
            \STATE Sample $\mathbf{\hat{Z}}_T \sim \mathcal{N}(0,1)$
            \STATE $\mathbf{D}=\operatorname{CPEN_{S2}}(\operatorname{PixelUnshuffle}(I_{LQ}))$ (paper Eq.~(\textbf{\textcolor{blue}{12}}))
            \FOR{$t=T$ to $1$ }
            \STATE  $\mathbf{\hat{Z}}_{t-1}=\frac{1}{\sqrt{\alpha_t}}\left(\mathbf{\hat{Z}}_t-\epsilon_{\theta}(\operatorname{Concat}(\mathbf{\hat{Z}}_t,t,\mathbf{D})) \frac{1-\alpha_t}{\sqrt{1-\bar{\alpha}_t}}\right)$ (paper Eq.~(\textbf{\textcolor{blue}{11}}))               
            \ENDFOR
            \STATE $\mathbf{\hat{Z}}=\mathbf{\hat{Z}}_{0}$
            \STATE $\hat{I}_{HQ} = \operatorname{DIRformer}(I_{LQ},\mathbf{\hat{Z}})$
   
		\STATE Output restored HQ images $\hat{I}_{HQ}$.
	\end{algorithmic}
\end{algorithm*}

\vspace{-1mm}
\section{More Training Details on Inpainting}
\vspace{-1mm}
 We train our DiffIR for inpainting using the same loss functions of LaMa~\cite{LaMa}, which further introduce multiple perceptual losses and adversarial loss to the basic $\mathcal{L}_{1}$ loss. 
 
For our experiments on image-inpainting in the paper Sec.~\textbf{\textcolor{blue}{5.2}}, we used the code of LaMa~\cite{LaMa} to generate synthetic masks. In training, we adopt the Adam optimizer with learning rates $0.0002$ and $0.0001$ for DiffIR and discriminator networks, respectively. All models are trained for 1M iterations with a batch size of 30. In addition, we use random crops of size $256\times256$ to train DiffIR on Places and CelebA-HQ. In testing, we use a fixed
set of 2k validation and 30k testing samples from CelebA-HQ~\cite{celeba} and Places~\cite{places2}. Moreover, we validate DiffIR$_{S2}$ on crops of size $512\times512$ and $256\times256$ on Places and CelebA-HQ validation datasets, respectively.

\vspace{-1mm}
\section{More Training Details on SR}
\vspace{-1mm}
Compared with DIRformer for other IR tasks, we add a $\times4$ upsampling network~\cite{ESRGAN} at the end of DIRformer for super-resolution (SR). We train our DiffIR for SR using the same loss functions of ESRGAN~\cite{ESRGAN}, which further introduce perceptual loss and adversarial loss to the basic $\mathcal{L}_{1}$ loss.

We train DiffIR for 1M iterations with a batch size of $64$. In addition, we use random crops of size $256\times256$ to train DiffIR on DIV2K~\cite{DIV2K}
(800 images) and Flickr2K~\cite{Flickr2K} (2650 images) datasets for
4$\times$ super-resolution. We train our DiffIR using Adam optimizer with learning rates $0.0002$ and $0.0001$ for DiffIR and discriminator networks, respectively.

\vspace{-1mm}
\section{More Training Details on deblurring}
\vspace{-1mm}
Following previous works in single image motion deblurring~\cite{MIMO-Unet, MPRNet,restormer}, we train our DiffIR only using $\mathcal{L}_{1}$ loss for fair comparisons. We train DiffIR for 300K iterations with the initial learning rate $2\time10^{-4}$ gradually reduced to $1\time10^{-6}$ with the cosine annealing~\cite{cosine}. Following previous work~\cite{restormer}, we progressively increase patch size and decrease batch size. Specifically, we start training with patch size $128\times128$
and batch size $64$. The patch size and batch size pairs are updated to $[(1602,40), (1922,32), (2562,16), (3202,8), (3842,8)]$ at iterations $[92K, 156K, 204K, 240K, 276K]$.

\vspace{-1mm}
\section{More Visual Comparisons on Inpainting}
\begin{figure*}[t]
    \setlength{\fsdurthree}{0mm}
    \Huge
    \centering
   \resizebox{1\linewidth}{!}{
            \begin{adjustbox}{valign=t}
                \begin{tabular}{cccccc}
                    \includegraphics[width=\widthscalethree \textwidth]{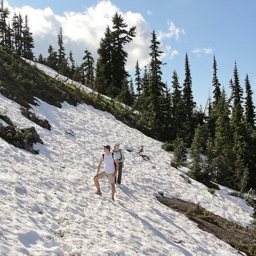} \hspace{\fsdurthree} &
                    \includegraphics[width=\widthscalethree  \textwidth]{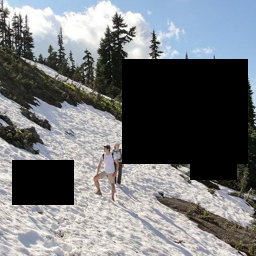} \hspace{\fsdurthree} &
                    \includegraphics[width=\widthscalethree  \textwidth]{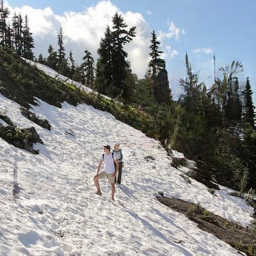} \hspace{\fsdurthree} &
                    \includegraphics[width=\widthscalethree  \textwidth]{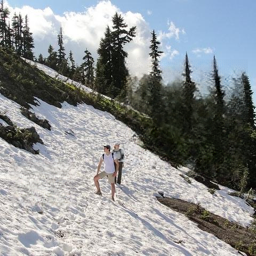} 
                    \hspace{\fsdurthree} &
                    \includegraphics[width=\widthscalethree  \textwidth]{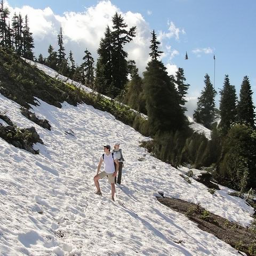} \hspace{\fsdurthree} &
                    \includegraphics[width=\widthscalethree  \textwidth]{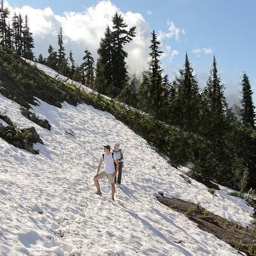} \hspace{\fsdurthree}  
                    \\
                    \includegraphics[width=\widthscalethree  \textwidth]{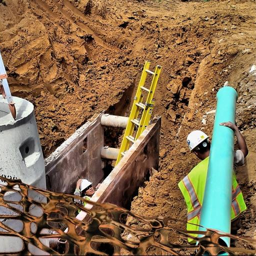} \hspace{\fsdurthree} &
                    \includegraphics[width=\widthscalethree  \textwidth]{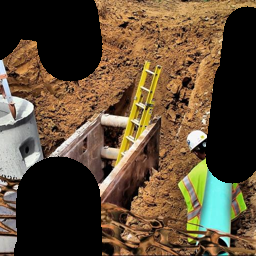} \hspace{\fsdurthree} &
                    \includegraphics[width=\widthscalethree  \textwidth]{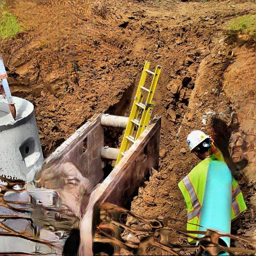} \hspace{\fsdurthree} &
                    \includegraphics[width=\widthscalethree  \textwidth]{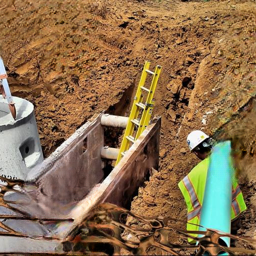} 
                    \hspace{\fsdurthree} &
                    \includegraphics[width=\widthscalethree  \textwidth]{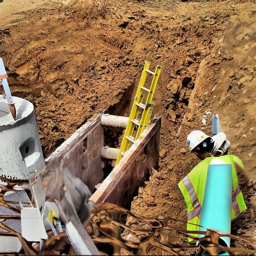} \hspace{\fsdurthree} &
                    \includegraphics[width=\widthscalethree  \textwidth]{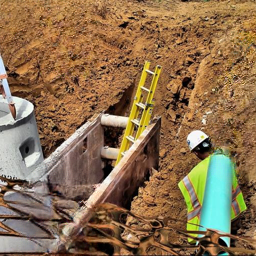} \hspace{\fsdurthree}  
                    \\
                    \includegraphics[width=\widthscalethree  \textwidth]{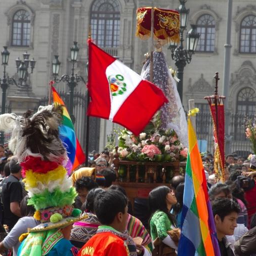} \hspace{\fsdurthree} &
                    \includegraphics[width=\widthscalethree  \textwidth]{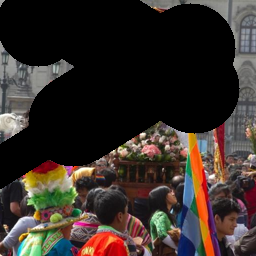} \hspace{\fsdurthree} &
                    \includegraphics[width=\widthscalethree  \textwidth]{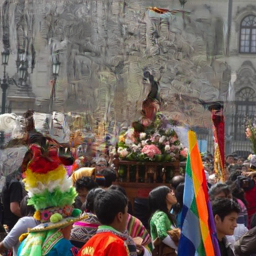} \hspace{\fsdurthree} &
                    \includegraphics[width=\widthscalethree  \textwidth]{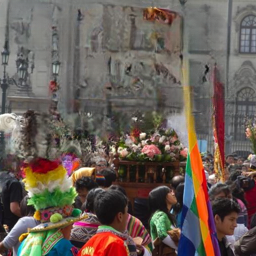} 
                    \hspace{\fsdurthree} &
                    \includegraphics[width=\widthscalethree  \textwidth]{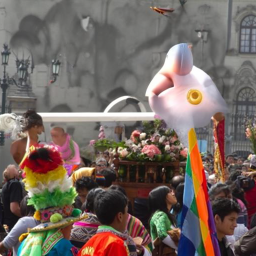} \hspace{\fsdurthree} &
                    \includegraphics[width=\widthscalethree  \textwidth]{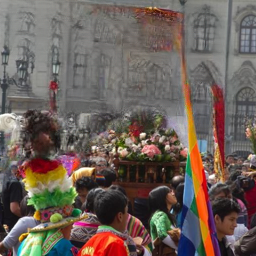} \hspace{\fsdurthree}  
                    \\
                    \includegraphics[width=\widthscalethree  \textwidth]{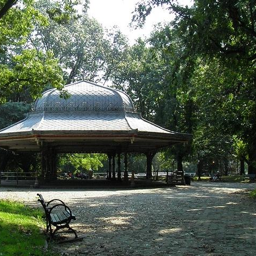} \hspace{\fsdurthree} &
                    \includegraphics[width=\widthscalethree  \textwidth]{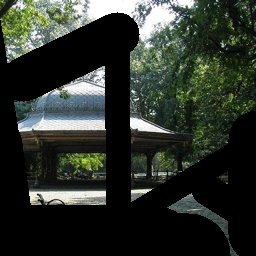} \hspace{\fsdurthree} &
                    \includegraphics[width=\widthscalethree  \textwidth]{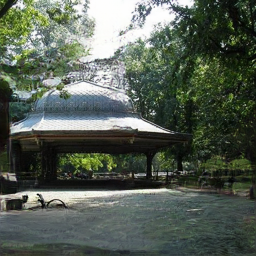} \hspace{\fsdurthree} &
                    \includegraphics[width=\widthscalethree  \textwidth]{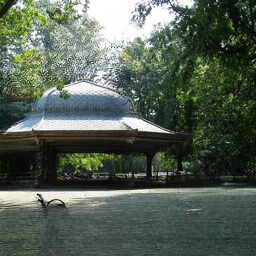} 
                    \hspace{\fsdurthree} &
                    \includegraphics[width=\widthscalethree  \textwidth]{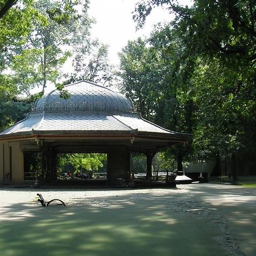} \hspace{\fsdurthree} &
                    \includegraphics[width=\widthscalethree  \textwidth]{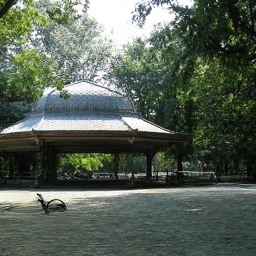} \hspace{\fsdurthree}  
                    \\
                    \includegraphics[width=\widthscalethree  \textwidth]{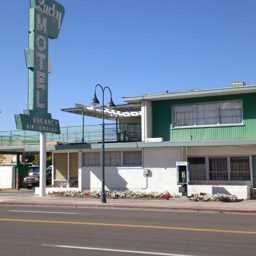} \hspace{\fsdurthree} &
                    \includegraphics[width=\widthscalethree  \textwidth]{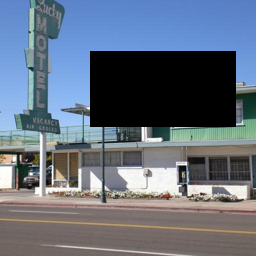} \hspace{\fsdurthree} &
                    \includegraphics[width=\widthscalethree  \textwidth]{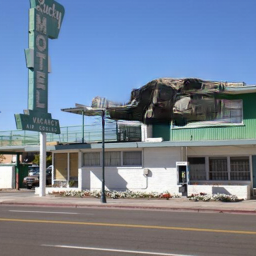} \hspace{\fsdurthree} &
                    \includegraphics[width=\widthscalethree  \textwidth]{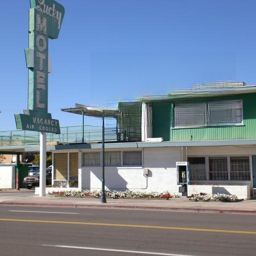} 
                    \hspace{\fsdurthree} &
                    \includegraphics[width=\widthscalethree  \textwidth]{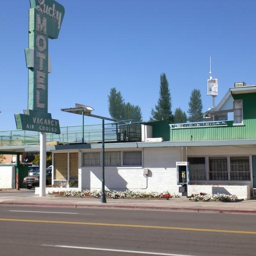} \hspace{\fsdurthree} &
                    \includegraphics[width=\widthscalethree  \textwidth]{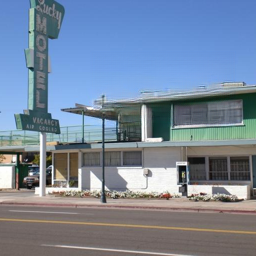} \hspace{\fsdurthree}  
                    \\
                    \includegraphics[width=\widthscalethree  \textwidth]{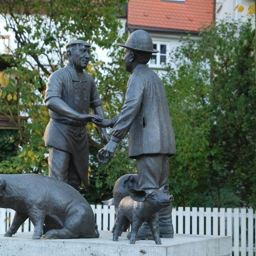} \hspace{\fsdurthree} &
                    \includegraphics[width=\widthscalethree  \textwidth]{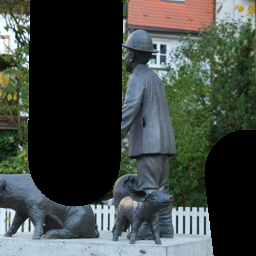} \hspace{\fsdurthree} &
                    \includegraphics[width=\widthscalethree  \textwidth]{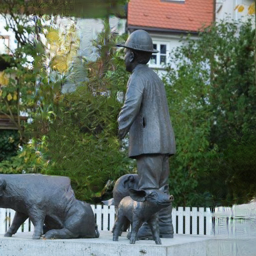} \hspace{\fsdurthree} &
                    \includegraphics[width=\widthscalethree  \textwidth]{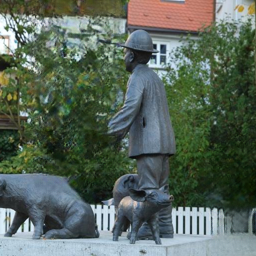} 
                    \hspace{\fsdurthree} &
                    \includegraphics[width=\widthscalethree  \textwidth]{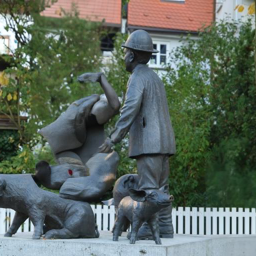} \hspace{\fsdurthree} &
                    \includegraphics[width=\widthscalethree  \textwidth]{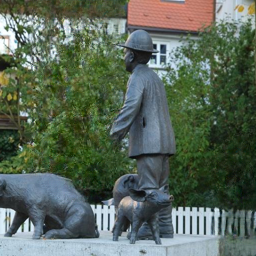} \hspace{\fsdurthree}  
                    \\
                    \includegraphics[width=\widthscalethree  \textwidth]{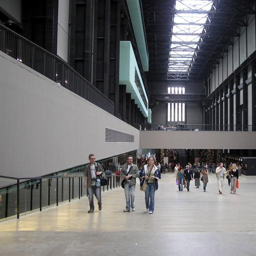} \hspace{\fsdurthree} &
                    \includegraphics[width=\widthscalethree  \textwidth]{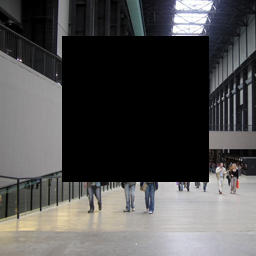} \hspace{\fsdurthree} &
                    \includegraphics[width=\widthscalethree  \textwidth]{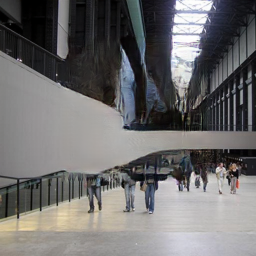} \hspace{\fsdurthree} &
                    \includegraphics[width=\widthscalethree  \textwidth]{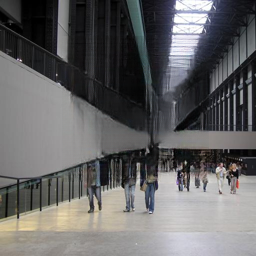} 
                    \hspace{\fsdurthree} &
                    \includegraphics[width=\widthscalethree  \textwidth]{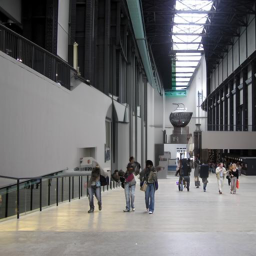} \hspace{\fsdurthree} &
                    \includegraphics[width=\widthscalethree  \textwidth]{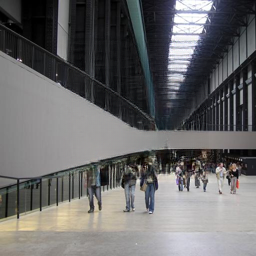} \hspace{\fsdurthree}  
                    \\
                    \includegraphics[width=\widthscalethree  \textwidth]{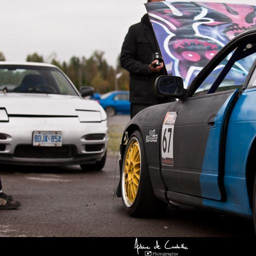} \hspace{\fsdurthree} &
                    \includegraphics[width=\widthscalethree  \textwidth]{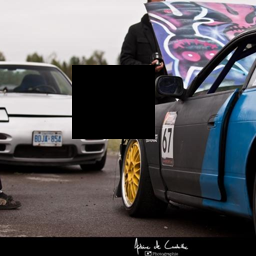} \hspace{\fsdurthree} &
                    \includegraphics[width=\widthscalethree  \textwidth]{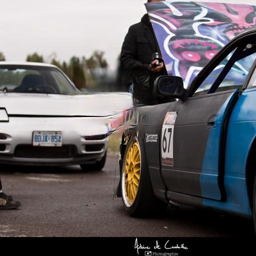} \hspace{\fsdurthree} &
                    \includegraphics[width=\widthscalethree  \textwidth]{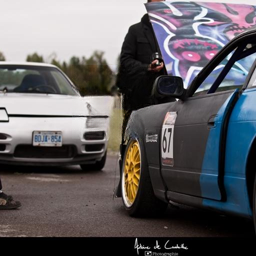} 
                    \hspace{\fsdurthree} &
                    \includegraphics[width=\widthscalethree  \textwidth]{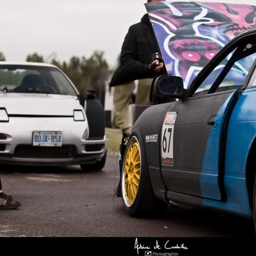} \hspace{\fsdurthree} &
                    \includegraphics[width=\widthscalethree  \textwidth]{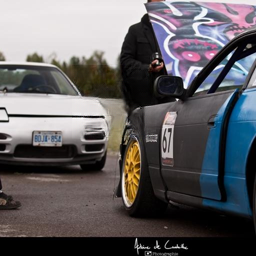} \hspace{\fsdurthree}  
                    \\
                    HQ \hspace{\fsdurthree} &
                    \makecell{LQ} \hspace{\fsdurthree} &
                    \makecell{ICT~\cite{ICT}} \hspace{\fsdurthree} &
                    LaMa~\cite{LaMa} \hspace{\fsdurthree} &
                    RePaint~\cite{repaint} \hspace{\fsdurthree} &
                    \makecell{DiffIR$_{S2}$ (Ours)} \hspace{\fsdurthree} 
                \end{tabular}
            \end{adjustbox}

    }
    \caption{More visual comparisons of \textbf{ inpainting} methods. Zoom-in for better details.}

    \label{fig:sup_inpainting_show}
\end{figure*}

\vspace{-1mm}
In this section, we provide more qualitative comparisons between our DiffIR$_{S2}$ and SOTA inpainting
methods (ICT~\cite{ICT}, LaMa~\cite{LaMa}, and RePaint~\cite{repaint}). The results are shown in Fig~\ref{fig:sup_inpainting_show}. We can observe that our DiffIR$_{S2}$ can produce more realistic and reasonable structures and details than other competitive inpainting methods. 

\section{More Visual Comparisons on SR}
\begin{figure*}[t]

    \setlength{\fsdurthree}{0mm}
    \LARGE
    \centering
   \resizebox{0.96\linewidth}{!}{
        \begin{tabular}{cc}
            \begin{adjustbox}{valign=t}
                \Large
                \begin{tabular}{c}
                    \includegraphics[height=0.44\textwidth]{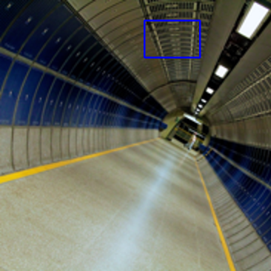} 
                \end{tabular}
                
            \end{adjustbox}
            
            \begin{adjustbox}{valign=t}
                \begin{tabular}{cccc}
                    \includegraphics[width=\widthscalefive \textwidth]{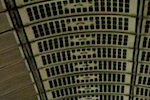} \hspace{\fsdurthree} &
                    \includegraphics[width=\widthscalefive \textwidth]{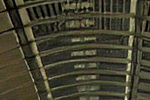} \hspace{\fsdurthree} &
                    \includegraphics[width=\widthscalefive \textwidth]{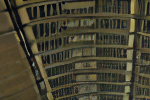} \hspace{\fsdurthree} 
                    \\
                    HQ \hspace{\fsdurthree} &
                    \makecell{BebyGAN~\cite{BebyGAN}} \hspace{\fsdurthree} &
                    \makecell{LDM~\cite{LDM}} \hspace{\fsdurthree} 
                    \\
                    \includegraphics[width=\widthscalefive \textwidth]{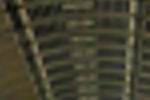} 
                    \hspace{\fsdurthree} &
                    \includegraphics[width=\widthscalefive \textwidth]{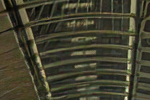} \hspace{\fsdurthree} &
                    \includegraphics[width=\widthscalefive \textwidth]{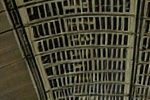} \hspace{\fsdurthree}  
                    \\ 
                    LQ \hspace{\fsdurthree} &
                    USRGAN~\cite{USRGAN} \hspace{\fsdurthree} &
                    \makecell{DiffIR$_{S2}$ (Ours)} \hspace{\fsdurthree} 
                \end{tabular}
            \end{adjustbox}
            \\
            \begin{adjustbox}{valign=t}
                \Large
                \begin{tabular}{c}
                    \includegraphics[height=0.44\textwidth]{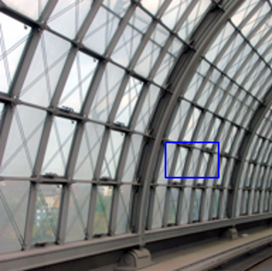} 
                \end{tabular}
                
            \end{adjustbox}
            
            \begin{adjustbox}{valign=t}
                \begin{tabular}{cccc}
                    \includegraphics[width=\widthscalefive \textwidth]{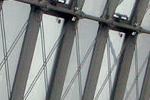} \hspace{\fsdurthree} &
                    \includegraphics[width=\widthscalefive \textwidth]{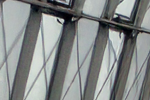} \hspace{\fsdurthree} &
                    \includegraphics[width=\widthscalefive \textwidth]{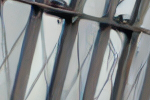} \hspace{\fsdurthree} 
                    \\
                    HQ \hspace{\fsdurthree} &
                    \makecell{BebyGAN~\cite{BebyGAN}} \hspace{\fsdurthree} &
                    \makecell{LDM~\cite{LDM}} \hspace{\fsdurthree} 
                    \\
                    \includegraphics[width=\widthscalefive \textwidth]{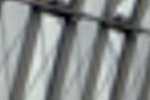} 
                    \hspace{\fsdurthree} &
                    \includegraphics[width=\widthscalefive \textwidth]{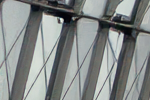} \hspace{\fsdurthree} &
                    \includegraphics[width=\widthscalefive \textwidth]{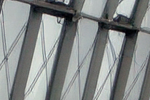} \hspace{\fsdurthree}  
                    \\ 
                    LQ \hspace{\fsdurthree} &
                    USRGAN~\cite{USRGAN} \hspace{\fsdurthree} &
                    \makecell{DiffIR$_{S2}$ (Ours)} \hspace{\fsdurthree} 
                \end{tabular}
            \end{adjustbox}
            \\
            \begin{adjustbox}{valign=t}
                \Large
                \begin{tabular}{c}
                    \includegraphics[height=0.44\textwidth]{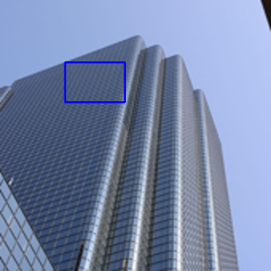} 
                \end{tabular}
                
            \end{adjustbox}
            
            \begin{adjustbox}{valign=t}
                \begin{tabular}{cccc}
                    \includegraphics[width=\widthscalefive \textwidth]{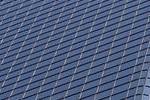} \hspace{\fsdurthree} &
                    \includegraphics[width=\widthscalefive \textwidth]{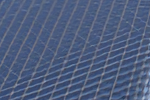} \hspace{\fsdurthree} &
                    \includegraphics[width=\widthscalefive \textwidth]{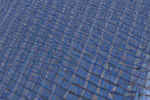} \hspace{\fsdurthree} 
                    \\
                    HQ \hspace{\fsdurthree} &
                    \makecell{BebyGAN~\cite{BebyGAN}} \hspace{\fsdurthree} &
                    \makecell{LDM~\cite{LDM}} \hspace{\fsdurthree} 
                    \\
                    \includegraphics[width=\widthscalefive \textwidth]{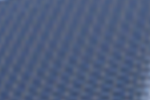} 
                    \hspace{\fsdurthree} &
                    \includegraphics[width=\widthscalefive \textwidth]{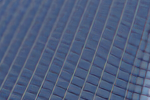} \hspace{\fsdurthree} &
                    \includegraphics[width=\widthscalefive \textwidth]{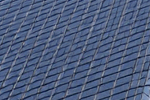} \hspace{\fsdurthree}  
                    \\ 
                    LQ \hspace{\fsdurthree} &
                    USRGAN~\cite{USRGAN} \hspace{\fsdurthree} &
                    \makecell{DiffIR$_{S2}$ (Ours)} \hspace{\fsdurthree} 
                \end{tabular}
            \end{adjustbox}
            \\
            \begin{adjustbox}{valign=t}
                \Large
                \begin{tabular}{c}
                    \includegraphics[height=0.44\textwidth]{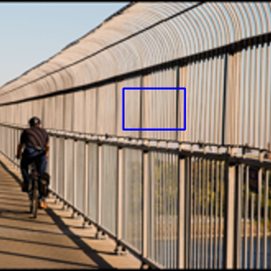} 
                \end{tabular}
                
            \end{adjustbox}
            
            \begin{adjustbox}{valign=t}
                \begin{tabular}{cccc}
                    \includegraphics[width=\widthscalefive \textwidth]{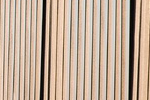} \hspace{\fsdurthree} &
                    \includegraphics[width=\widthscalefive \textwidth]{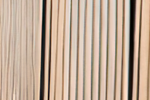} \hspace{\fsdurthree} &
                    \includegraphics[width=\widthscalefive \textwidth]{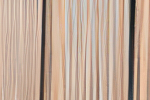} \hspace{\fsdurthree} 
                    \\
                    HQ \hspace{\fsdurthree} &
                    \makecell{BebyGAN~\cite{BebyGAN}} \hspace{\fsdurthree} &
                    \makecell{LDM~\cite{LDM}} \hspace{\fsdurthree} 
                    \\
                    \includegraphics[width=\widthscalefive \textwidth]{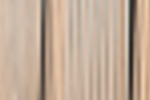} 
                    \hspace{\fsdurthree} &
                    \includegraphics[width=\widthscalefive \textwidth]{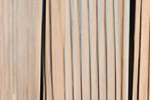} \hspace{\fsdurthree} &
                    \includegraphics[width=\widthscalefive \textwidth]{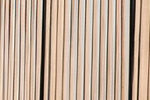} \hspace{\fsdurthree}  
                    \\ 
                    LQ \hspace{\fsdurthree} &
                    USRGAN~\cite{USRGAN} \hspace{\fsdurthree} &
                    \makecell{DiffIR$_{S2}$ (Ours)} \hspace{\fsdurthree} 
                \end{tabular}
            \end{adjustbox}

        \end{tabular}
    }
    \caption{ Visual comparison of 4$\times$ \textbf{image super-resolution} methods. Zoom-in for better details.}
    \label{fig:sup_SR_showv1}
\end{figure*}

\begin{figure*}[t]

    \setlength{\fsdurthree}{0mm}
    \LARGE
    \centering
   \resizebox{0.96\linewidth}{!}{
        \begin{tabular}{cc}
            \begin{adjustbox}{valign=t}
                \Large
                \begin{tabular}{c}
                    \includegraphics[height=0.44\textwidth]{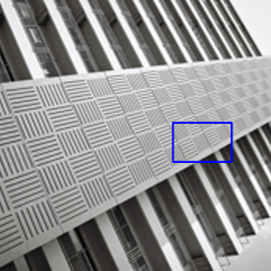} 
                \end{tabular}
                
            \end{adjustbox}
            
            \begin{adjustbox}{valign=t}
                \begin{tabular}{cccc}
                    \includegraphics[width=\widthscalefive \textwidth]{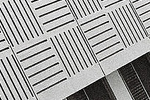} \hspace{\fsdurthree} &
                    \includegraphics[width=\widthscalefive \textwidth]{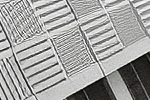} \hspace{\fsdurthree} &
                    \includegraphics[width=\widthscalefive \textwidth]{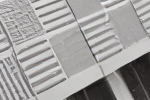} \hspace{\fsdurthree} 
                    \\
                    HQ \hspace{\fsdurthree} &
                    \makecell{BebyGAN~\cite{BebyGAN}} \hspace{\fsdurthree} &
                    \makecell{LDM~\cite{LDM}} \hspace{\fsdurthree} 
                    \\
                    \includegraphics[width=\widthscalefive \textwidth]{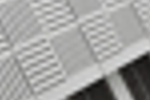} 
                    \hspace{\fsdurthree} &
                    \includegraphics[width=\widthscalefive \textwidth]{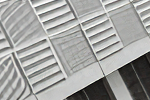} \hspace{\fsdurthree} &
                    \includegraphics[width=\widthscalefive \textwidth]{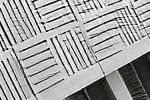} \hspace{\fsdurthree}  
                    \\ 
                    LQ \hspace{\fsdurthree} &
                    USRGAN~\cite{USRGAN} \hspace{\fsdurthree} &
                    \makecell{DiffIR$_{S2}$ (Ours)} \hspace{\fsdurthree} 
                \end{tabular}
            \end{adjustbox}
            \\
            \begin{adjustbox}{valign=t}
                \Large
                \begin{tabular}{c}
                    \includegraphics[height=0.44\textwidth]{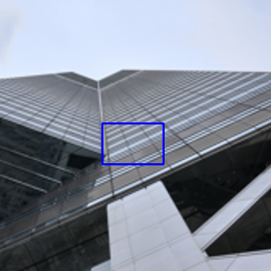} 
                \end{tabular}
                
            \end{adjustbox}
            
            \begin{adjustbox}{valign=t}
                \begin{tabular}{cccc}
                    \includegraphics[width=\widthscalefive \textwidth]{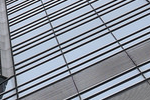} \hspace{\fsdurthree} &
                    \includegraphics[width=\widthscalefive \textwidth]{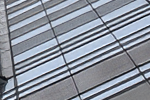} \hspace{\fsdurthree} &
                    \includegraphics[width=\widthscalefive \textwidth]{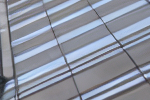} \hspace{\fsdurthree} 
                    \\
                    HQ \hspace{\fsdurthree} &
                    \makecell{BebyGAN~\cite{BebyGAN}} \hspace{\fsdurthree} &
                    \makecell{LDM~\cite{LDM}} \hspace{\fsdurthree} 
                    \\
                    \includegraphics[width=\widthscalefive \textwidth]{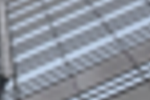} 
                    \hspace{\fsdurthree} &
                    \includegraphics[width=\widthscalefive \textwidth]{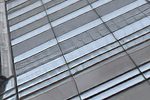} \hspace{\fsdurthree} &
                    \includegraphics[width=\widthscalefive \textwidth]{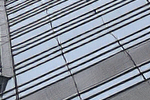} \hspace{\fsdurthree}  
                    \\ 
                    LQ \hspace{\fsdurthree} &
                    USRGAN~\cite{USRGAN} \hspace{\fsdurthree} &
                    \makecell{DiffIR$_{S2}$ (Ours)} \hspace{\fsdurthree} 
                \end{tabular}
            \end{adjustbox}
            \\
            \begin{adjustbox}{valign=t}
                \Large
                \begin{tabular}{c}
                    \includegraphics[height=0.44\textwidth]{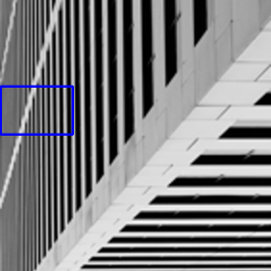} 
                \end{tabular}
                
            \end{adjustbox}
            
            \begin{adjustbox}{valign=t}
                \begin{tabular}{cccc}
                    \includegraphics[width=\widthscalefive \textwidth]{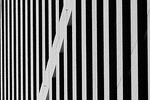} \hspace{\fsdurthree} &
                    \includegraphics[width=\widthscalefive \textwidth]{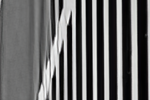} \hspace{\fsdurthree} &
                    \includegraphics[width=\widthscalefive \textwidth]{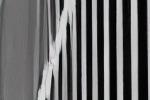} \hspace{\fsdurthree} 
                    \\
                    HQ \hspace{\fsdurthree} &
                    \makecell{BebyGAN~\cite{BebyGAN}} \hspace{\fsdurthree} &
                    \makecell{LDM~\cite{LDM}} \hspace{\fsdurthree} 
                    \\
                    \includegraphics[width=\widthscalefive \textwidth]{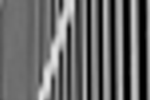} 
                    \hspace{\fsdurthree} &
                    \includegraphics[width=\widthscalefive \textwidth]{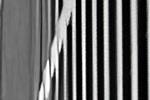} \hspace{\fsdurthree} &
                    \includegraphics[width=\widthscalefive \textwidth]{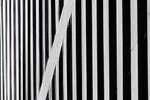} \hspace{\fsdurthree}  
                    \\ 
                    LQ \hspace{\fsdurthree} &
                    USRGAN~\cite{USRGAN} \hspace{\fsdurthree} &
                    \makecell{DiffIR$_{S2}$ (Ours)} \hspace{\fsdurthree} 
                \end{tabular}
            \end{adjustbox}
            \\
            \begin{adjustbox}{valign=t}
                \Large
                \begin{tabular}{c}
                    \includegraphics[height=0.44\textwidth]{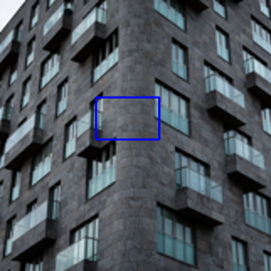} 
                \end{tabular}
                
            \end{adjustbox}
            
            \begin{adjustbox}{valign=t}
                \begin{tabular}{cccc}
                    \includegraphics[width=\widthscalefive \textwidth]{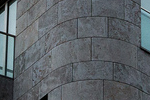} \hspace{\fsdurthree} &
                    \includegraphics[width=\widthscalefive \textwidth]{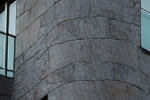} \hspace{\fsdurthree} &
                    \includegraphics[width=\widthscalefive \textwidth]{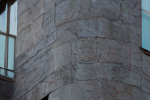} \hspace{\fsdurthree} 
                    \\
                    HQ \hspace{\fsdurthree} &
                    \makecell{BebyGAN~\cite{BebyGAN}} \hspace{\fsdurthree} &
                    \makecell{LDM~\cite{LDM}} \hspace{\fsdurthree} 
                    \\
                    \includegraphics[width=\widthscalefive \textwidth]{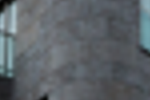} 
                    \hspace{\fsdurthree} &
                    \includegraphics[width=\widthscalefive \textwidth]{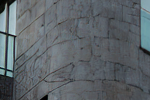} \hspace{\fsdurthree} &
                    \includegraphics[width=\widthscalefive \textwidth]{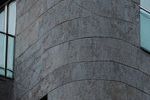} \hspace{\fsdurthree}  
                    \\ 
                    LQ \hspace{\fsdurthree} &
                    USRGAN~\cite{USRGAN} \hspace{\fsdurthree} &
                    \makecell{DiffIR$_{S2}$ (Ours)} \hspace{\fsdurthree} 
                \end{tabular}
            \end{adjustbox}

        \end{tabular}
    }
    \caption{ Visual comparison of 4$\times$ \textbf{image super-resolution} methods. Zoom-in for better details.}
    \label{fig:sup_SR_showv2}
\end{figure*}

In this section, we provide more qualitative comparisons between our DiffIR$_{S2}$ and SOTA GAN-based SR methods. The results are shown in Figs~\ref{fig:sup_SR_showv1} and~\ref{fig:sup_SR_showv2}.  Our DiffIR$_{S2}$  achieves the best visual quality containing more realistic details.

\section{More Visual Comparisons on Deblurring}
\begin{figure*}[t]
    \setlength{\fsdurthree}{0mm}
    \LARGE
    \centering
   \resizebox{0.96\linewidth}{!}{
        \begin{tabular}{cc}
            \begin{adjustbox}{valign=t}
                \Large
                \begin{tabular}{c}
                    \includegraphics[height=0.44\textwidth]{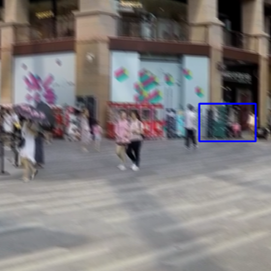} 
                \end{tabular}
                
            \end{adjustbox}
            
            \begin{adjustbox}{valign=t}
                \begin{tabular}{cccc}
                    \includegraphics[width=\widthscalefive \textwidth]{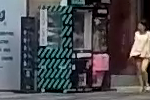} \hspace{\fsdurthree} &
                    \includegraphics[width=\widthscalefive \textwidth]{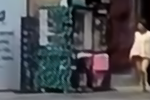} \hspace{\fsdurthree} &
                    \includegraphics[width=\widthscalefive \textwidth]{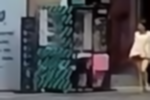} \hspace{\fsdurthree} 
                    \\
                    HQ \hspace{\fsdurthree} &
                    \makecell{MT-RNN~\cite{MT-RNN}} \hspace{\fsdurthree} &
                    \makecell{Restormer~\cite{restormer}} \hspace{\fsdurthree} 
                    \\
                    \includegraphics[width=\widthscalefive \textwidth]{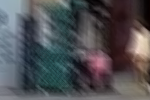} 
                    \hspace{\fsdurthree} &
                    \includegraphics[width=\widthscalefive \textwidth]{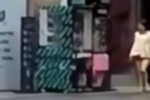} \hspace{\fsdurthree} &
                    \includegraphics[width=\widthscalefive \textwidth]{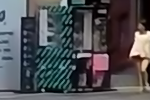} \hspace{\fsdurthree}  
                    \\ 
                    LQ \hspace{\fsdurthree} &
                    MPRNet~\cite{MPRNet} \hspace{\fsdurthree} &
                    \makecell{DiffIR$_{S2}$ (Ours)} \hspace{\fsdurthree} 
                \end{tabular}
            \end{adjustbox}
            \\
            \begin{adjustbox}{valign=t}
                \Large
                \begin{tabular}{c}
                    \includegraphics[height=0.44\textwidth]{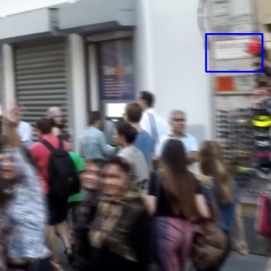} 
                \end{tabular}
                
            \end{adjustbox}
            
            \begin{adjustbox}{valign=t}
                \begin{tabular}{cccc}
                    \includegraphics[width=\widthscalefive \textwidth]{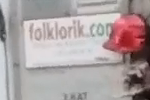} \hspace{\fsdurthree} &
                    \includegraphics[width=\widthscalefive \textwidth]{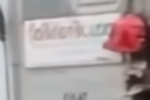} \hspace{\fsdurthree} &
                    \includegraphics[width=\widthscalefive \textwidth]{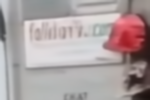} \hspace{\fsdurthree} 
                    \\
                    HQ \hspace{\fsdurthree} &
                    \makecell{MT-RNN~\cite{MT-RNN}} \hspace{\fsdurthree} &
                    \makecell{Restormer~\cite{restormer}} \hspace{\fsdurthree} 
                    \\
                    \includegraphics[width=\widthscalefive \textwidth]{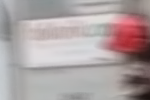} 
                    \hspace{\fsdurthree} &
                    \includegraphics[width=\widthscalefive \textwidth]{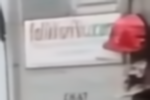} \hspace{\fsdurthree} &
                    \includegraphics[width=\widthscalefive \textwidth]{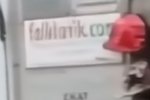} \hspace{\fsdurthree}  
                    \\ 
                    LQ \hspace{\fsdurthree} &
                    MPRNet~\cite{MPRNet} \hspace{\fsdurthree} &
                    \makecell{DiffIR$_{S2}$ (Ours)} \hspace{\fsdurthree} 
                \end{tabular}
            \end{adjustbox}
            \\
            \begin{adjustbox}{valign=t}
                \Large
                \begin{tabular}{c}
                    \includegraphics[height=0.44\textwidth]{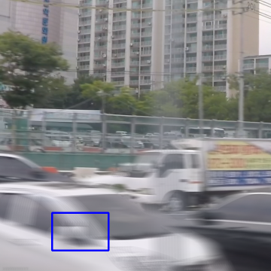} 
                \end{tabular}
                
            \end{adjustbox}
            
            \begin{adjustbox}{valign=t}
                \begin{tabular}{cccc}
                    \includegraphics[width=\widthscalefive \textwidth]{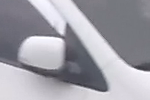} \hspace{\fsdurthree} &
                    \includegraphics[width=\widthscalefive \textwidth]{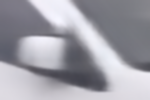} \hspace{\fsdurthree} &
                    \includegraphics[width=\widthscalefive \textwidth]{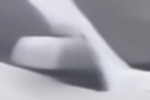} \hspace{\fsdurthree} 
                    \\
                    HQ \hspace{\fsdurthree} &
                    \makecell{MT-RNN~\cite{MT-RNN}} \hspace{\fsdurthree} &
                    \makecell{Restormer~\cite{restormer}} \hspace{\fsdurthree} 
                    \\
                    \includegraphics[width=\widthscalefive \textwidth]{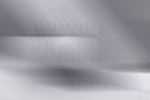} 
                    \hspace{\fsdurthree} &
                    \includegraphics[width=\widthscalefive \textwidth]{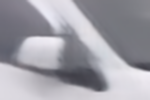} \hspace{\fsdurthree} &
                    \includegraphics[width=\widthscalefive \textwidth]{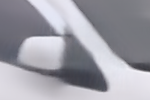} \hspace{\fsdurthree}  
                    \\ 
                    LQ \hspace{\fsdurthree} &
                    MPRNet~\cite{MPRNet} \hspace{\fsdurthree} &
                    \makecell{DiffIR$_{S2}$ (Ours)} \hspace{\fsdurthree} 
                \end{tabular}
            \end{adjustbox}
            \\
            \begin{adjustbox}{valign=t}
                \Large
                \begin{tabular}{c}
                    \includegraphics[height=0.44\textwidth]{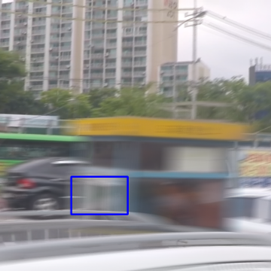} 
                \end{tabular}
                
            \end{adjustbox}
            
            \begin{adjustbox}{valign=t}
                \begin{tabular}{cccc}
                    \includegraphics[width=\widthscalefive \textwidth]{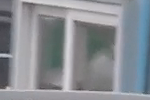} \hspace{\fsdurthree} &
                    \includegraphics[width=\widthscalefive \textwidth]{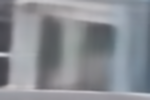} \hspace{\fsdurthree} &
                    \includegraphics[width=\widthscalefive \textwidth]{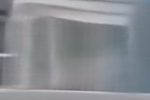} \hspace{\fsdurthree} 
                    \\
                    HQ \hspace{\fsdurthree} &
                    \makecell{MT-RNN~\cite{MT-RNN}} \hspace{\fsdurthree} &
                    \makecell{Restormer~\cite{restormer}} \hspace{\fsdurthree} 
                    \\
                    \includegraphics[width=\widthscalefive \textwidth]{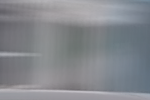} 
                    \hspace{\fsdurthree} &
                    \includegraphics[width=\widthscalefive \textwidth]{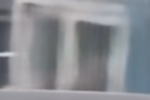} \hspace{\fsdurthree} &
                    \includegraphics[width=\widthscalefive \textwidth]{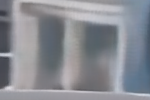} \hspace{\fsdurthree}  
                    \\ 
                    LQ \hspace{\fsdurthree} &
                    MPRNet~\cite{MPRNet} \hspace{\fsdurthree} &
                    \makecell{DiffIR$_{S2}$ (Ours)} \hspace{\fsdurthree} 
                \end{tabular}
            \end{adjustbox}
        \end{tabular}
    }
    \caption{Visual comparison of \textbf{single image motion deblurring} methods. Zoom-in for better details.}
    \label{fig:sup_deblur_show}
    \vspace{-3mm}
\end{figure*}

In this section, we provide more qualitative comparisons between our DiffIR$_{S2}$ and SOTA image motion deblurring methods. The results are shown in Fig~\ref{fig:sup_deblur_show}.  Our DiffIR$_{S2}$ has the best visual quality containing more realistic details close to corresponding HQ images.

\end{document}